\documentclass[journal]{IEEEtran}

\usepackage{cite}
\usepackage[pdftex]{graphicx}
\usepackage{amsmath, amssymb}
\usepackage{mathtools}
\usepackage{algorithmic,algorithm}
\usepackage{array}
\usepackage{enumerate}
\usepackage[caption=false,font=footnotesize]{subfig}

\usepackage{enumitem}
\usepackage{adjustbox}
\usepackage{booktabs} % for professional tables
\usepackage[table]{xcolor}
\usepackage{xcolor}
\usepackage{color,soul}
\usepackage{url}
\definecolor{Gray}{gray}{0.9}

\usepackage{changes}
\definechangesauthor[name={Laurent}, color=red]{LNA}
\definechangesauthor[name={Aditya}, color=blue]{AC}

\begin{document}

%\title{Hyperspectral Image Classification using Watershed Classifiers}
\title{Triplet-Watershed for Hyperspectral Image Classification}

\author{Aditya~Challa,
        Sravan~Danda,~\IEEEmembership{Member,~IEEE,}
        B.S.Daya~Sagar,~\IEEEmembership{Senior~Member,~IEEE,}
        and~Laurent~Najman,~\IEEEmembership{Senior~Member,~IEEE}% <-this % stops a space
\thanks{Aditya Challa is with the Department of Computer Science and Information Systems, BITS Pilani K K Birla Goa Campus,
NH-17B, Zuarinagar, Goa 403726 e-mail:aditya.challa.20@gmail.com.}% <-this % stops a space
\thanks{Sravan~Danda is with APPCAIR, Department of Computer Science and Information Systems, BITS Pilani K K Birla Goa Campus,
NH-17B, Zuarinagar, Goa 403726. email:sravan8809@gmail.com }%
\thanks{B.~S.~Daya~Sagar is with Systems Science and Informatics Unit, Indian Statistical Institute, Bengaluru, Karnataka, 560059 email: bsdsagar@yahoo.co.uk}% <-this % stops a space
\thanks{Laurent Najman is with Universit\'e Gustave Eiffel, LIGM, Equipe A3SI, ESIEE, France. email:laurent.najman@esiee.fr}% <-this % stops a space
}

% \markboth{Journal of \LaTeX\ Class Files,~Vol.~14, No.~8, August~2015}%
% {Shell \MakeLowercase{\textit{et al.}}: Bare Demo of IEEEtran.cls for IEEE Journals}


\maketitle

\begin{abstract}
  Hyperspectral images (HSI) consist of rich spatial and spectral information, which can potentially be used for several applications. However, noise, band correlations and high dimensionality restrict the applicability of such data. This is recently addressed using creative deep learning network architectures such as ResNet, SSRN, and A2S2K. However, the last layer, i.e the classification layer, remains unchanged and is taken to be the softmax classifier. In this article, we propose to use a watershed classifier. Watershed classifier extends the watershed operator from Mathematical Morphology for classification. In its vanilla form, the watershed classifier does not have any trainable parameters. In this article, we propose a novel approach to train deep learning networks to obtain representations suitable for the watershed classifier. The watershed classifier exploits the connectivity patterns, a characteristic of HSI datasets, for better inference. We show that exploiting such characteristics allows the Triplet-Watershed to achieve state-of-art results in supervised and semi-supervised contexts. These results are validated on Indianpines (IP), University of Pavia (UP), Kennedy Space Center (KSC)  and University of Houston (UH) datasets, relying on simple convnet architecture using a quarter of parameters compared to previous state-of-the-art networks. 
  
  The source code for reproducing the experiments and supplementary material (high resolution images) is available at \url{https://github.com/ac20/TripletWatershed_Code}.
\end{abstract}

\begin{IEEEkeywords}
Hyperspectral Imaging, Watershed, Triplet Loss, Deep Learning, Classification
\end{IEEEkeywords}

\section{Introduction}

\IEEEPARstart{H}{yperspectral} imaging has several applications ranging across different domains \cite{IEEE/journals/grsm/Ghamsi2017}. It has seen applications in earth observations \cite{DBLP:journals/tgrs/MelganiB04}{, and} land cover classification \cite{DBLP:journals/prl/GislasonBS06} etc. Hyperspectral datasets have rich information both spatially and spectrally. However, spectral and spatial correlations make a lot of such information redundant. One can obtain efficient representations using techniques such as band selection \cite{DBLP:journals/tgrs/CaiLC20,IEEE:journals/grsl/SKRoy20} subspace learning \cite{hong2020joint,SPRINGER:conferences/eccv/hong2018} multi-modal learning \cite{IEEE/journals/tgrs/hong2020More} low-rank representation \cite{RemoteSensing/journals/rs/Gao2017}.

Large number of bands, spatial and spectral feature correlations and curse of dimensionality make Hyperspectral image classification challenging. Conventional approaches use hand crafted features with techniques such as scale-invariant feature transform (SIFT) \cite{ELSEVIER/journals/prl/Li2019} sparse representation \cite{ELSEVIER/journals/pr/Shao2018} principal component analysis \cite{IEEE/journals/grsl/Licciardi2012} independent component analysis \cite{IEEE/journals/tgrs/Villa2011}. {Classic approaches to classification such as support vector machines (SVM) \cite{DBLP:journals/tgrs/MelganiB04}, neural networks \cite{IEEE/journals/tgrs/Zhong2012} and logistic regression \cite{IEEE/journals/tgrs/Li2010} aimed at exploiting the spectral signatures alone. Using spatial features have been extremely useful to obtain better representations and higher classification accuracies \cite{IEEE/journals/grsm/Ghamisi2018,IEEE/journals/tgrs/He2018,IEEE/journals/tgrs/Li2019}, which the classic approaches ignore}. Multiple kernel learning \cite{IEEE/journals/grsl/Camps-Valls2006,ELSEVIER/journals/pr/Fauvel2012,IEEE/journals/tgrs/Fang2015} use hand-designed kernels to exploit the spectral-spatial interactions. Deep learning approaches, especially CNNs, have been adapted to exploit the spectral-spatial information. \cite{IEEE/journals/tgrs/Chen2016} proposes a 3D-CNN feature extractor to obtain combined spectral-spatial features. \cite{IEEE/journals/tgrs/Yang2017} adapts CNN to a two-branch architecture to extract joint spectral-spatial features. \cite{IEEE/journals/tgrs/Hamida2018} used 3D volumes to extract spectral-spatial features, which may be improved using multi-scale approaches \cite{IEEE/conferences/icip/He2017}. Spectral-spatial residual network (SSRN) proposed in \cite{IEEE/journals/tgrs/Zhong2018} uses residual networks to remove the declining accuracy phenomenon. Residual Spectral–Spatial Attention Networks (RSSAN) \cite{IEEE/journals/tgrs/Zhu2021} exploit the concept of attention to improve on SSRNs. \cite{IEEE/journals/tgrs/Roy2020} proposes Attention-Based Adaptive Spectral-Spatial Kernel Residual networks (A2S2K) by exploiting adaptive kernels. \cite{IEEE/journals/tgrs/Hong2020} uses graph convolution networks and \cite{IEEE/journals/tgrs/Paoletti2019} uses capsule networks. Most of these approaches tackle the problem of Hyperspectral image classification by considering novel architectures. Another prominent direction of research focusses on using unlabelled data for improving classification accuracies, referred to as semi-supervised learning. In \cite{ELSEVIER:journal/isprsjprs/Hong2019,IEEE:journal/tgrs/Hong2019} the authors use hyperspectral data for improving inference on multispectral data. In \cite{IEEE/journals/tgrs/Hong2020} the authors propose a semi-supervised approach to exploit multi-modal data for better inference. Graph Convolution Networks (GCN) have also been used to obtain state-of-art results on hyperspectral classification as evidenced by S2GCN\cite{IEEE:journal/grsl/Qin2019} and DC-GCN (Dual Clustering GCN)\cite{ARXIV:Zeng2020}. {Other approaches include  local constraint-based sparse manifold hypergraph learning (LC-SMHL) \cite{IEEE:journals/jstars/YuleHY21}, self-adaptive manifold discriminant analysis (SAMDA) \cite{ELSIVIER:Journals/PR/HuangLHDS20}, DLPNet \cite{ELSIVIER:Journals/nn/LiHDS20} and adaptive residual convolutional neural network (ARCNN) \cite{IEEE:journals/jstars/HongCL20}.}

In this article, we take a different route to propose a novel classifier based on the watershed operator. Watershed operator from Mathematical Morphology \cite{DBLP:journals/pami/VincentS91,conference/misc/Serge1993} has been widely used for image segmentation, and, in particular, for Hyperspectral images \cite{noyel2007morphological,tarabalka2010segmentation}. In \cite{tarabalka2010segmentation}, the authors combine (by majority voting) several watersheds computed on gradients of different bands. They observe that {\em class-specific accuracies were improved by using the spatial information in the classification for almost all the classes}, a result that we are going to confirm in the present paper. { To our knowledge, watersheds have not been used in conjunction with current state-of-art neural networks in the context of hyperspectral images. We propose a novel approach to achieve this in the current article.}

In \cite{DBLP:journals/pami/CoustyBNC09} the watershed operator is adapted to edge-weighted graphs. It is shown there that the watershed is closely related to the minimum spanning tree (MST) of the graph. Watersheds have also been highly successful as a post-processing tool for image segmentation \cite{DBLP:conf/nips/TuragaBHDS09,IEEE/journals/pami/Maninis2018,DBLP:conf/iccv/WolfSKH17}. In \cite{IEEE:journals/pami/Funke2019} the authors learn a representation suitable for MST-based classification. In \cite{IEEE:journals/pami/Wolf2020} the authors learn a representation suitable to mutex-watershed, a different version of the watershed. 

Departing from images, in our previous work \cite{DBLP:journals/spl/ChallaDSN19} we have proposed to use the watershed operator as a generic classifier. We showed that it obtains a \emph{maximum margin partition} similar to the support vector machine. We further showed that ensemble watersheds obtain comparable performance to other classifiers such as random forests. In this article we propose a novel approach, simple and efficient, called {\em Triplet-Watershed} to learn representations (also known as embeddings) suitable for the watershed classifier.

\begin{table}[!t]
  \caption{\label{table:1}Overall Accuracy (OA) vs Number of Parameters. Observe that the proposed method has very less number of parameters but outperforms the current state-of-the-art approaches. IP indicates Indian Pines dataset. UP denotes University of Pavia dataset and KSC indicates the Kennedy Space Centre dataset.}
  \centering
  \begin{tabular}{lcccc}\toprule
   & $\#$ params & IP & UP & KSC\\
  \midrule
  A2S2K\cite{IEEE/journals/tgrs/Roy2020} & 370.7K & 98.66& 99.85& 99.34\\
  \rowcolor{Gray}SSRN\cite{IEEE/journals/tgrs/Zhong2018} & 364.1K & 98.38& 99.77&99.29\\
  ENL-FCN\cite{IEEE:journals/tgrs/Shen2020} & 113.9K & 96.15&99.76 &99.25\\
  \rowcolor{Gray}ResNet34\cite{DBLP:conf/cvpr/HeZRS16} & 21.9M & 92.44&97.38 &79.73\\
  \midrule
  \textbf{Triplet-Watershed} & \textbf{87.6K} & \textbf{99.57} & \textbf{99.98} & \textbf{99.72}\\
  \bottomrule
  \end{tabular}
  \end{table}

\emph{Why watershed classifier?} Previous work on hyperspectral image classification, as discussed above, establish that one must use both spatial and spectral aspects to obtain good classifiers. They achieve this with creative approaches to design neural networks such as adaptive kernels, attention mechanism, etc. However, most of these still use conventional softmax classifier. The watershed classifier naturally uses spatial information for inference. Thus, it allows us to use simpler networks for representation. Table \ref{table:1} shows the overall accuracy scores obtained by our approach and other state of art methods. It also shows the number of parameters used. Observe that Triplet-Watershed parameters are just 25$\%$ of those of the current state-of-art (A2S2K) approach.

{ The main contributions of this article are the following.
\begin{enumerate}[label=(\roman*)]
  \item We propose a novel approach, namely {\em the Triplet-Watershed}, to learn a representation suitable to the watershed classifier. This representation is referred to as \emph{watershed representations} in the rest of the article.
  \item The Triplet-Watershed achieves state-of-art results on the hyperspectral datasets with very simple networks, using much fewer parameters than the previous state-of-the-art approaches as described in table \ref{table:1}.
  \item The same Triplet-Watershed approach can be used for both supervised and semi-supervised tasks without any modification, still leading to state-of-the-art results compared to previous approaches.
  \item The framework used here to obtain representations is not restricted to watershed classifiers. It can be extended to use with other classifiers such as random forest or $k$-nearest neighbours as well, although watershed results outperform other classifiers on our datasets.
  \item The main insight of our paper is that enforcing spatial connectivity (achieved thanks to the watershed classifier) during the training is a relevant constraint for hyperspectral classification.
\end{enumerate}}

\emph{Overview:} Section~\ref{sec:watershed classifier} reviews the watershed classifier and the required terminology for the rest of the article. In section~\ref{sec:watershedreprensentations} we design the neural net (NN) and the training procedure to learn watershed representations. Section~\ref{sec:empirical analysis} provides empirical analysis. 

\section{Watershed Classifier}
\label{sec:watershed classifier}
The watershed classifier is defined on an edge-weighted graph. We follow the exposition as given in \cite{DBLP:journals/spl/ChallaDSN19}.  $G = (V, E, W)$ denotes the edge-weighted graph. Here $V$ denotes the set of vertices, $E$ denotes the set of edges which is a subset of $V \times V$ and $W : E \to \mathbb{R}^{+}$ denotes the edge weight assigned to each edge. We assume that the edge weights are all positive in this article.

\begin{figure}[!t]
  \centering
  \includegraphics[width=0.6\linewidth]{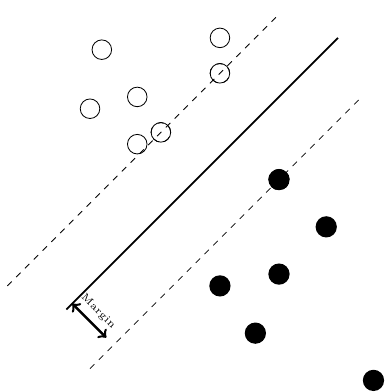}
  \caption{Illustration of maximum margin for support vector machines (SVM)\cite{DBLP:journals/spl/ChallaDSN19}. The key observation is - The margin is defined as the minimum distance between the training point labelled $0$ and what would be labelled $1$ after classification. And vice versa. The aim of the (linear) SVM classifier is to obtain a decision boundary that maximizes the margin. This can be extended to obtain a maximum-margin partition on an edge-weighted graph. Using \eqref{eq:max margin}, a solution of this is provided by the watershed classifier.}
  \label{fig:1}
  \end{figure}

The (two-class) classification problem on the edge-weighted graph is stated as  - Let $X_0, X_1 \subset V$ denote the labelled subset of vertices labelled $0$ and $1$ respectively. Classification problem requires a partition of $V = M_0 \cup M_1$ with $M_0 \cap M_1 = \emptyset$. With an additional constraint that $X_0 \subset M_0$ and $X_1 \subset M_1$. Here $M_0$ denotes all the vertices labelled $0$ after classification and $M_1$ denotes all the vertices labelled $1$. We also assume there exists a dissimilarity measure $\rho(x,y)$ between two vertices $x, y \in V$. This measure extends to subsets as
\begin{equation}
  \rho(X, Y) = \min_{x \in X, y \in Y} \rho(x,y)
\end{equation}
where $X, Y$ are arbitrary subsets of $V$. Observe that there exist several partitions of $V = M_0 \cup M_1$ which satisfy the above conditions. Of these partitions, we only use the \emph{Maximum Margin Partitions}, i.e the partitions which maximize
\begin{equation}
  \min \{\rho(X_0, M_1), \rho(X_1, M_0)\}
  \label{eq:max margin}
\end{equation}
This follows from the maximum margin principle of support vector machines (SVM). From figure \ref{fig:1}, a key observation can be made - The margin for the SVM is the minimum distance between training point labelled $0$ and what would be labelled $1$ after classification. And vice versa. Linear SVM tries to obtain the boundary to maximize this margin. This can be extended to the edge-weighted graphs using \eqref{eq:max margin}.

\begin{figure}[!t]
\centering
\subfloat[]{\includegraphics[width=0.3\linewidth]{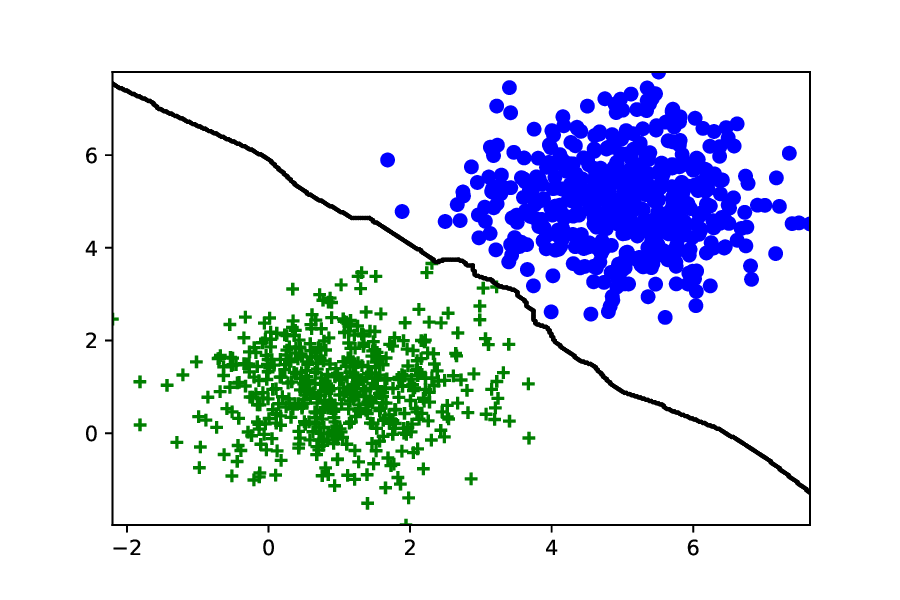}%
\label{fig:2a}}
\hfil
\subfloat[]{\includegraphics[width=0.3\linewidth]{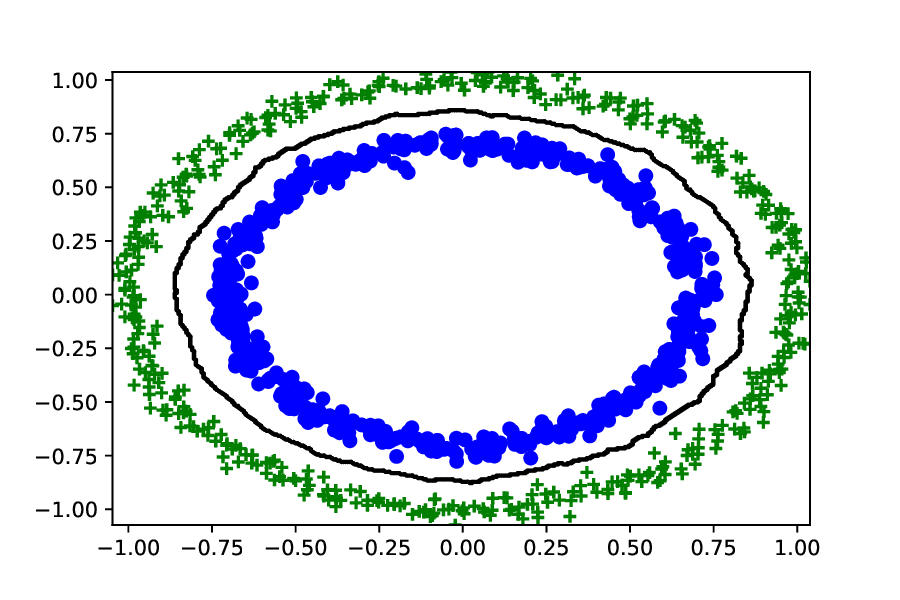}%
\label{fig:2b}}
\hfil
\subfloat[]{\includegraphics[width=0.3\linewidth]{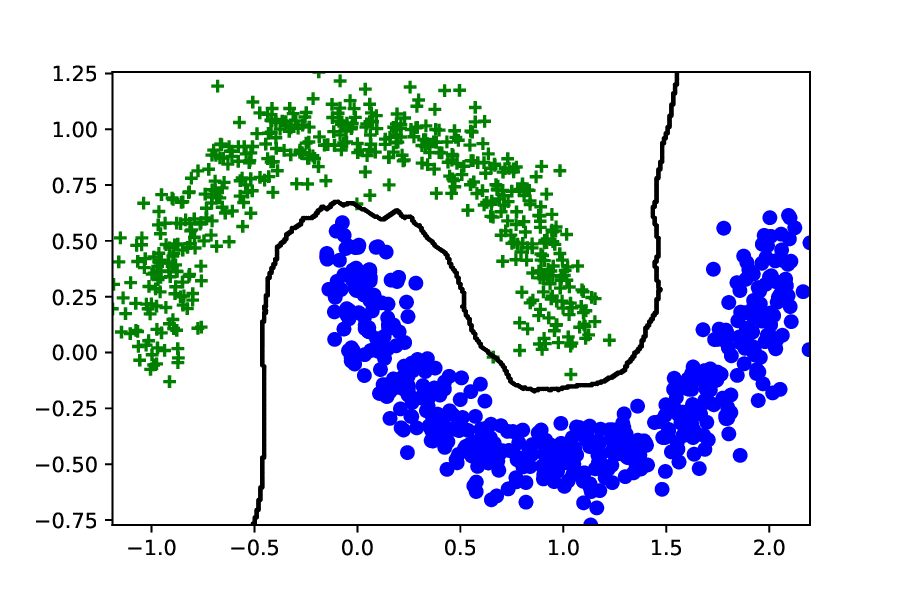}%
\label{fig:2c}}
\caption{Figure illustrating the watershed boundaries\cite{DBLP:journals/spl/ChallaDSN19}. Observe that in all these cases the boundary is in-between the classes. Also, it is in the middle of the zero density (no points exist) regions. This maximizes the margin between the boundaries and the classes. This is consistent with the maximum margin principle of SVM.}
\label{fig:2}
\end{figure}

\begin{figure}
  
  \centering
  \subfloat[Original Graph]{\includegraphics[width=0.33\linewidth]{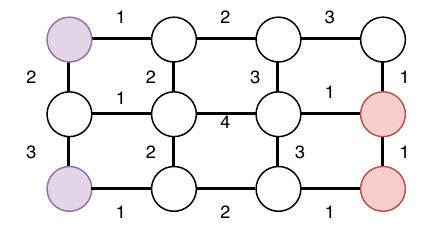}%
    \label{fig:R1_watershed1}}
  \hfil
  \subfloat[Intermediate Step]{\includegraphics[width=0.33\linewidth]{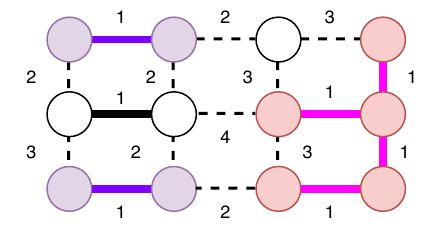}%
    \label{fig:R1_watershed2}}
  \hfil
  \subfloat[Watershed Labels]{\includegraphics[width=0.33\linewidth]{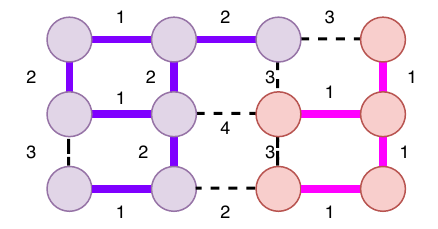}%
    \label{fig:R1_watershed3}}
  \caption{Illustrating the watershed classifier. Let (a) denote the edge-weighted. The two distinct colours indicate two different classes. No colour indicates that the vertex is not yet labelled. (b) denotes the graph obtained by adding edges with weight $1$. Each vertex is given a label accordingly. (c) denotes the graph obtained by adding the edges with weight $2$ and Propagating the labels. Observe that all the points are now labelled and hence the algorithm terminates. }
  \label{fig:R1_watershed}
\end{figure}

The \emph{Watershed Classifier} is defined by considering the dissimilarity measure to be 
\begin{equation}
  \rho(x,y) \coloneqq \rho_{max}(x,y) = \min_{\pi \in \Pi(x,y)} \max_{e \in \pi} W(e)
  \label{eq:pass value}
\end{equation}
where $\pi$ denotes a specific path between $x, y$. $\Pi$ denotes the set of all possible paths. $\rho_{max}$ is sometimes referred to as \emph{pass value}. 

If each edge-weight indicates the height of the corresponding edge, then $\rho_{max}(x,y)$ indicates the minimum height one has to climb to reach $y$ from $x$. When the points belong to a Euclidean space, the edge weight is given by Euclidean distance. That is, the edge weight indicates the distance between the points. Hence, $\rho_{max}(x,y)$ would indicate the minimum ``jump'' one has to make to reach $y$ from $x$. Thus, the boundaries (in cases where the classes are separable) would be along the low-density regions between classes. This is illustrated in Figure \ref{fig:2}. In all the cases, the boundary is between the classes such that we have the maximum margin. This is consistent with the maximum margin principle of SVM. 

{
\noindent
\textbf{Remark:} One can replace the pass value in \eqref{eq:pass value} with several other measures, leading to different classifiers. Detailed analysis of replacing pass value with other measures is out of scope for the present article and may be considered for future work. For instance, using the Image Foresting Transform (IFT) \cite{DBLP:journals/pami/FalcaoSL04} leads to a classifier similar to the one proposed in \cite{DBLP:conf/sibgrapi/AmorimFC14}. Few such techniques are discussed in \cite{DBLP:journals/spl/ChallaDSN19}.
}

Given the edge-weighted graph, the \textbf{Watershed algorithm} extends the Maximum Margin Partition principle to several classes and obtains the labels using the \textsc{UnionFind} data structure. This is described in algorithm \ref{alg:watershed}.

\begin{algorithm}[H]
  \caption{Watershed clustering algorithm \cite{DBLP:journals/spl/ChallaDSN19}}
  \label{alg:watershed}
\begin{algorithmic}[1]
  \renewcommand{\algorithmicrequire}{\textbf{Input:}}
  \renewcommand{\algorithmicensure}{\textbf{Output:}}
  \REQUIRE edge-weighted graph $G = (V, E, W)$. A subset of labelled points $V_l \subset V$.
  \ENSURE  Labels for each of the vertices $L$
    \STATE Sort the edges $E$ in increasing order w.r.t $W$.
    \STATE Initialize the union-find data structure \textsc{UF},
  \FOR {$e = (e_x,e_y)$ in sorted edge set $E$}
    \IF {both $e_x$ and $e_y$ are labelled}
      \STATE do nothing
    \ELSE
      \STATE \textsc{UF}.union$(e_x, e_y)$
      \STATE Assign same label for $e_x$ and $e_y$.
    \ENDIF
  \ENDFOR
  \STATE Label each vertex of the connected component using labels $V_l$. \label{line:labelling}
  \RETURN Labels of the vertices.
\end{algorithmic} 
\end{algorithm}

Observe that step (10) is possible since each connected component would have exactly one unique label. One can see that watershed clustering is a semi-supervised algorithm, in the sense that it propagates the known labels to points with unknown label. 

{
To illustrate the watershed classifier consider the simple edge-weighted graph in figure \ref{fig:R1_watershed1}. The two distinct colours indicate two classes. No colour indicates that the vertex is not yet labelled. In the first step, the least edge-weight is $1$. Adding all these edges (thick edges in figure \ref{fig:R1_watershed2}) gives 4 distinct components. Each component is labelled according to the label present within the component. In case there exists no label, then label assignment is not yet carried out. We then add the edges with weight $2$, and label the points accordingly. Observe that there are no more unlabelled points and hence the algorithm terminates.
}

In practice, it has been observed that ensemble techniques improve the robustness of watershed classifier. This is achieved using only a subset of labelled points and only a subset of features and taking the weighted average. Details can be found in \cite{DBLP:journals/spl/ChallaDSN19}. We refer to these two approaches as {\em single watershed classifier} and {\em ensemble-watershed classifier}.

\section{Learning Representations for the Watershed Classifier}
\label{sec:watershedreprensentations}

The previous section described how one can obtain the labels using the watershed classifier. In \cite{DBLP:journals/spl/ChallaDSN19}, it was shown that this compares reasonably well to other classifiers such as SVM, random forests, etc. However, observe that this classifier has \emph{no trainable parameters}. In this section, we develop an approach to train a neural network for learning representations suitable to the watershed classifier.

\begin{figure}[!t]
  \centering
  \includegraphics[width=1.0\linewidth]{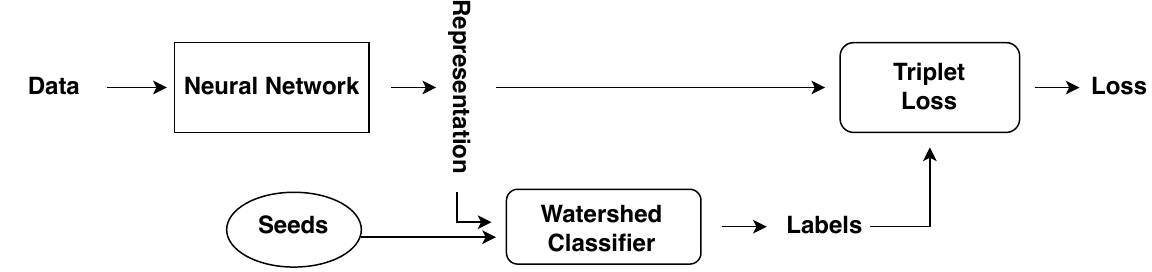}
  \caption{Schematic of learning representations for the watershed classifier. Using a generic neural network we obtain the representation for the dataset. These representations are fed into the watershed classifier to obtain the labels using the seeds. Using the labels and the representation, we use triplet loss to compute the loss and also for obtaining the parameters for the neural network. Observe that the watershed classifier needs to be computed at every epoch.}
  \label{fig:3b}
  \end{figure}

A key observation is - Watershed classifier reduces the distances within each component and increases the distance across components. This leads to the schematic in figure \ref{fig:3b}. First, we use a generic neural network to obtain the representations for the dataset. These representations, along with a subset of labelled points, are used with the watershed classifier to obtain the labels. Using these labels, we obtain a metric-learning loss to decide if two pixels are either in the same component (same label) of the watershed or in two different components (different label). More precisely, we use triplet loss \cite{DBLP:journals/corr/HofferA14,DBLP:conf/nips/SchultzJ03} to learn the watershed representation. For training, this cost is minimized using standard autograd packages such as pytorch.

\emph{Why schematic in figure \ref{fig:3b} learns watershed representations?} Triplet loss function uses $\{\text{(anchor, postive, negative)}\}$ triplets for computation of the cost. It compares an anchor-input to a positive-input and a negative-input. The distance from the anchor-input to the positive-input is minimized, and the distance from the anchor-input to the negative-input is maximized using the cost
\begin{equation}
    \min \{d(\text{anchor}, \text{positive}) - d(\text{anchor}, \text{negative}) + \alpha\}_{+}
\end{equation}
where $\{* \}_{+}$ denotes the function $\max\{0, *\}$. By enforcing the order of distances, triplet loss models embed in the way that a pair of samples with the same label are smaller in distance than those with different labels. When watershed labels are used to obtain $\{\text{(anchor, postive, negative)}\}$ triplets, this leads to representations that are compatible with the watershed classifier. 

{\noindent
  \textbf{Remark (Supervised vs Semi-Supervised)} : Recall that the watershed classifier uses a subset of training points (referred to as seeds) to obtain the labels of other training points. These labels are then used to the train the network with triplet loss. However, in the case of semi-supervised learning, unlabelled data is also available at train time. These points can be labelled and be used to train the network. In this article we use the semi-supervised approach, randomly choosing some seeds for the watershed classifier that iteratively propagates their labels to their most resembling neighbours, obtaining the connected components. Hence, the combination of watershed clustering and triplet loss ensures that points with the most resembling representations are indeed clustered together, in the same connected component. 
}

\subsection*{Training Dynamics}

To summarize the entire training procedure of Triplet-Watershed, at each epoch 
\begin{enumerate}
  \item Obtain the representations for all the points using the neural network.
  \item We consider a randomly chosen subset of labelled points as seeds
  \item Propagate the labels to all points using the watershed classifier
  \item Use the watershed labels to generate $\{\text{(anchor, positive, negative)}\}$ triplets 
  \item Use the triplet loss to train the neural network. 
\end{enumerate}
Few obvious questions follow - (a) When would the training converge? (b) What is the steady-state obtained?

Note that the training would converge when there would be no further improvement in the triplet-loss. At this stage, the out-of-box score\footnote{Accuracy on the training data excluding the seeds} of the watershed classifier would not improve as well. This implies that - all pairs of points with the same labels and within the same component have similar representation. Hence, we obtain 100\% out-of-box accuracy\footnote{Here we assume that there exists at least one seed per component} with watershed classifier. 

{\noindent
\textbf{Remark (Overfitting):} Traditional machine learning advices against reaching 100\% training accuracy as the models might be overfitting. However, recent deep learning trends point to the contrary. Several deep learning models can indeed fit random data with 100\% accuracy \cite{DBLP:conf/iclr/ZhangBHRV17}. It is still an open question to understand the generalization ability of these models. However, few observations point to the \emph{inductive bias} \cite{DBLP:journals/corr/abs-1806-01261} as the reason behind good generalization. In our case, the inductive bias is dictated by the graph constructed from the data.
}

Also, note that during training we use a single watershed classifier. While, at inference, we use an ensemble-watershed classifier. This ensures robustness during inference. 

{\noindent

\textbf{Remark (Complexity):} Two main steps can be identified in the above procedure - (i) Obtaining a representation of the points and (ii) Propagating the labels using watershed. Time complexity for obtaining the representation is dictated by matrix multiplications with the neural network. This can easily be parallelized using GPU. Empirical study of the time taken for this is discussed in the following section. Table \ref{table:R1_8} shows the actual time taken for both training and evaluation. Propagation of labels is done using binary partition trees and can be performed in quasi-linear time \cite{DBLP:conf/ismm/NajmanCP13}. We use the routines available at \cite{DBLP:journals/softx/PerretCCGKN19} for implementation.
}

\section{Empirical Analysis}
\label{sec:empirical analysis}

In this section, we explore the application of the watershed classifier to the hyperspectral image classification task. We use the standard evaluation metrics for comparison: 
\begin{enumerate}[label=(\roman*)]
    \item Overall Accuracy (OA): it measures the overall accuracy across all samples, not considering the class imbalance. 
    \item Average Accuracy (AA): it measures the average accuracy across the classes and
    \item Kappa Coefficient ($\kappa$): it measures how well the estimates and groundtruth labels correspond, taking into account agreement by random chance. 
\end{enumerate}
Four datasets are used for comparison.
\begin{itemize}
  \item \textbf{Indian Pines (IP) :} Gathered by the Airborne Visible/Infrared Imaging Spectrometer (AVIRIS \cite{ELSEVIER:journals/rse/Green1998}) sensor over the test site in North-western Indiana. This data set contains 224 spectral bands within a wavelength range of $0.4 \text{ to } 2.5 \times 10^{-6}$ meters. The 24 bands covering region of water absorption are removed. The image spatial dimension is $145\times 145$, and there are $16$ classes not all mutually exclusive. 
  \item \textbf{Kennedy Space Centre (KSC) :} The Kennedy Space Center (KSC) data set was gathered on March 23, 1996 by AVIRIS \cite{ELSEVIER:journals/rse/Green1998} with wavelengths ranging from $0.4 \text{ to } 2.5\times 10^{-6}$ meters. 176 spectral bands are used for analysis after removal of some low signal-to-noise ratio (SNR) bands and water absorption bands. 13 classes representing the various land cover types that occur in this environment are defined for the site.
  \item \textbf{University of Pavia (UP) :} Acquired by the ROSIS \cite{doi:conference/Kunkel1988} sensor during a flight campaign over Pavia, northern Italy. The number of spectral bands is 103 for Pavia University and is of size $610\times 610$ pixels. The ground truth identifies 9 classes. 
  \item { \textbf{University of Houston (UH) :} This dataset was acquired over the University of Houston campus and the neighbouring urban area. This dataset was captures with a spatial resolution of 2.5m and with 144 spectral bands in the 380 nm to 1050 nm region. This has $15$ groundtruth classes. The dataset can be obtained from \url{https://hyperspectral.ee.uh.edu/?page_id=459}\footnote{Accessed on 30 April 2021.}.}
\end{itemize}
We preprocess the datasets using principal component analysis (PCA) \cite{doi:journals/Karl1901} to obtain orthogonal components. We use $200$ principal components for IP, $176$ for KSC, $103$ for UP and $144$ for UH datasets. The train/test split is obtained randomly using $10\%$ for training and $90\%$ for testing. 

\emph{Graph Construction:} Note that the watershed classifier is defined on edge-weighted graphs. This is constructed as follows
\begin{itemize}
  \item The set of vertices $V$ is taken to be the set of all the pixels in the dataset ignoring the $\{\text{labels}=0\}$ class. Since, these points do not have any groundtruth labels.
  \item The edge set $E$ is taken to be the union of 4-adjacency edges induced by the vertex set $V$ (on the image) and edges obtained by EMST (Euclidean Minimum Spanning Tree \cite{DBLP:conf/kdd/MarchRG10}) for Indianpines (IP), University of Pavia (UP) and Kennedy Space Centre (KSC), and K-Neighbour edges with k=50 for University of Houston (UH) dataset. The EMST and K-Neighbour edges are obtained by considering the top $32$ principal components.
  \item Given a representation obtained thanks to the neural network, the edge weights are computed using Euclidean distance. This representation (and hence the edge weights themselves) is updated at every epoch during training, while the edge set itself is never updated.
\end{itemize}

{
An illustration of the graph construction procedure is provided in appendix \ref{sec:appendix 1}.
}

\begin{figure}[!t]
  \centering
  \includegraphics[width=1.0\linewidth]{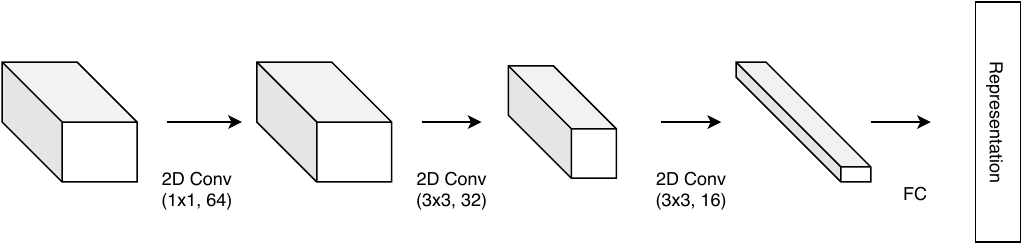}
  \caption{Neural Network architecture used for obtaining the representations. The architecture is composed of 3 convolution layers followed by a fully connected layer to get the representation. Batch normalization is performed before each layer for efficient training. The number of parameters is 87K.}
  \label{fig:3a}
\end{figure}

\begin{table*}
\caption{Overall Accuracy (OA), Average Accuracy (AA), and Kappa($\kappa$) values on Indianpines (IP) dataset using 10\% of samples for training.}
\label{table:2}
\centering
\begin{adjustbox}{max width=.95\linewidth}
\begin{tabular}{ccccccccccc}
    \toprule
        \phantom{abc} & \phantom{abc} & \phantom{abc} &\phantom{abc}& \multicolumn{3}{c}{Classic approaches} & \phantom{abc} & \multicolumn{3}{c}{Deep-Learning approaches}\\
         \cmidrule{5-7}  \cmidrule{9-11}
    Class& Train & Test &\phantom{abc}& RF\cite{IEEE:journals/tgrs/Ham2005} & SVM\cite{DBLP:journals/tgrs/MelganiB04} & Ensemble-Watershed\cite{DBLP:journals/spl/ChallaDSN19} &\phantom{abc}& SSRN\cite{IEEE/journals/tgrs/Zhong2018}& A2S2K\cite{IEEE/journals/tgrs/Roy2020} &Triplet-Watershed\\
    \midrule
                    1 & 4    & 42   &\phantom{abc}& 28.46 $\pm$ 0.061 & 51.22 $\pm$ 0.190 &  41.43 $\pm$ 0.2079 &\phantom{abc}& 57.78 $\pm$ 0.423 & 97.56 $\pm$ 0.034 & \textbf{100.00 $\pm$ 0.0000}   \\
    \rowcolor{Gray} 2 & 142  & 1286 &\phantom{abc}& 56.63 $\pm$ 0.024 & 81.22 $\pm$ 0.037 &  81.07 $\pm$ 0.0202 &\phantom{abc}& 98.37 $\pm$ 0.012 & \textbf{98.62 $\pm$ 0.010} & \textbf{98.62 $\pm$ 0.0151}   \\
                    3 & 83   & 747  &\phantom{abc}& 48.42 $\pm$ 0.013 & 65.82 $\pm$ 0.013 &  71.49 $\pm$ 0.0250 &\phantom{abc}& 97.47 $\pm$ 0.010 & 98.58 $\pm$ 0.006 & \textbf{100.00 $\pm$ 0.0000}   \\
    \rowcolor{Gray} 4 & 23   & 214  &\phantom{abc}& 33.49 $\pm$ 0.025 & 57.75 $\pm$ 0.041 &  45.70 $\pm$ 0.0327 &\phantom{abc}& 99.12 $\pm$ 0.010 & 98.29 $\pm$ 0.014 & \textbf{100.00 $\pm$ 0.0000}   \\
                    5 & 48   & 435  &\phantom{abc}& 85.21 $\pm$ 0.025 & 90.04 $\pm$ 0.014 &  92.78 $\pm$ 0.0286 &\phantom{abc}& 97.79 $\pm$ 0.013 & \textbf{99.02 $\pm$ 0.003} & {97.98 $\pm$ 0.0254}   \\
    \rowcolor{Gray} 6 & 73   & 657  &\phantom{abc}& 92.64 $\pm$ 0.027 & 96.25 $\pm$ 0.006 &  98.57 $\pm$ 0.0033 &\phantom{abc}& 98.50 $\pm$ 0.010 & 98.71 $\pm$ 0.010 & \textbf{99.97 $\pm$ 0.0006}   \\
                    7 & 2    & 26   &\phantom{abc}&  2.67 $\pm$ 0.038 & 73.33 $\pm$ 0.019 &  99.17 $\pm$ 0.0167 &\phantom{abc}& 66.67 $\pm$ 0.471 & 93.10 $\pm$ 0.097 & \textbf{100.00 $\pm$ 0.0000}   \\
    \rowcolor{Gray} 8 & 47   & 431  &\phantom{abc}& 97.67 $\pm$ 0.015 & 97.98 $\pm$ 0.006 &  98.14 $\pm$ 0.0075 &\phantom{abc}& 96.45 $\pm$ 0.029 & 98.83 $\pm$ 0.016 & \textbf{100.00 $\pm$ 0.0000}   \\
                    9 & 2    & 18   &\phantom{abc}&  9.26 $\pm$ 0.094 & 50.00 $\pm$ 0.045 &  37.50 $\pm$ 0.1854 &\phantom{abc}& 56.25 $\pm$ 0.418 & 74.26 $\pm$ 0.038 & \textbf{100.00 $\pm$ 0.0000}   \\
    \rowcolor{Gray} 10 & 97  & 875  &\phantom{abc}& 60.91 $\pm$ 0.047 & 73.87 $\pm$ 0.018 &  85.81 $\pm$ 0.0227 &\phantom{abc}& 98.33 $\pm$ 0.009 & 98.21 $\pm$ 0.016 & \textbf{99.75 $\pm$ 0.0040}   \\
                    11 & 245 & 2210 &\phantom{abc}& 87.88 $\pm$ 0.019 & 82.90 $\pm$ 0.012 &  86.68 $\pm$ 0.0105 &\phantom{abc}& 99.08 $\pm$ 0.005 & 99.09 $\pm$ 0.001 & \textbf{99.61 $\pm$ 0.0054}   \\
    \rowcolor{Gray} 12 & 59  & 534  &\phantom{abc}& 41.26 $\pm$ 0.030 & 74.91 $\pm$ 0.043 &  69.51 $\pm$ 0.0182 &\phantom{abc}& 98.46 $\pm$ 0.009 & 98.37 $\pm$ 0.013 & \textbf{99.89 $\pm$ 0.0022}   \\
                    13 & 20  & 185  &\phantom{abc}& 90.09 $\pm$ 0.040 & 96.94 $\pm$ 0.021 &  99.35 $\pm$ 0.0079 &\phantom{abc}& 100.0 $\pm$ 0.000 & 99.80 $\pm$ 0.002 & \textbf{100.00 $\pm$ 0.0000}   \\
    \rowcolor{Gray} 14 & 126 & 1139 &\phantom{abc}& 95.46 $\pm$ 0.014 & 93.82 $\pm$ 0.010 &  92.59 $\pm$ 0.0085 &\phantom{abc}& 98.63 $\pm$ 0.010 & 99.22 $\pm$ 0.007 & \textbf{100.00 $\pm$ 0.0000}   \\
                    15 & 38  & 348  &\phantom{abc}& 41.11 $\pm$ 0.029 & 60.42 $\pm$ 0.044 &  54.48 $\pm$ 0.0396 &\phantom{abc}& 99.24 $\pm$ 0.005 & 97.86 $\pm$ 0.013 & \textbf{100.00 $\pm$ 0.0000}   \\
    \rowcolor{Gray} 16 & 9   & 84   &\phantom{abc}& 79.37 $\pm$ 0.030 & 91.27 $\pm$ 0.054 &  79.29 $\pm$ 0.1163 &\phantom{abc}& 95.63 $\pm$ 0.062 & 95.93 $\pm$ 0.057 & \textbf{98.10 $\pm$ 0.0267}   \\
    \midrule
                    OA & 1018 & 9231 &\phantom{abc}& 72.98 $\pm$ 0.006  & 82.00 $\pm$ 0.006  & 83.75 $\pm$ 0.0076  &\phantom{abc}& 98.38 $\pm$ 0.004  & 98.66 $\pm$ 0.004& \textbf{99.57 $\pm$ 0.0026} \\
    \rowcolor{Gray} AA &      &      &\phantom{abc}& 59.41 $\pm$ 0.005  & 77.36 $\pm$ 0.019  & 77.10 $\pm$ 0.0228  &\phantom{abc}& 91.11 $\pm$ 0.080  & 96.59 $\pm$ 0.003& \textbf{99.62 $\pm$ 0.0029} \\
    $\kappa$ &&      &\phantom{abc}& 0.6862 $\pm$ 0.007 & 0.7941 $\pm$ 0.007 & 0.8143 $\pm$ 0.0086  &\phantom{abc}& 0.9815 $\pm$ 0.005 & 0.9848 $\pm$ 0.005& \textbf{0.9951 $\pm$ 0.0030} \\
    \bottomrule
    \end{tabular}
\end{adjustbox}
\end{table*}

\begin{table*}
\caption{Overall Accuracy (OA), Average Accuracy (AA), and Kappa($\kappa$) values on University of Pavia (UP) dataset using 10\% of samples for training.}
\label{table:3}
\centering
\begin{adjustbox}{max width=.95\linewidth}
\begin{tabular}{ccccccccccc}
    \toprule
        \phantom{abc} & \phantom{abc} & \phantom{abc} &\phantom{abc}& \multicolumn{3}{c}{Classic approaches} & \phantom{abc} & \multicolumn{3}{c}{Deep-Learning approaches}\\
                 \cmidrule{5-7}  \cmidrule{9-11}
    Class& Train & Test &\phantom{abc}& RF\cite{IEEE:journals/tgrs/Ham2005} & SVM\cite{DBLP:journals/tgrs/MelganiB04} & Ensemble-Watershed\cite{DBLP:journals/spl/ChallaDSN19} &\phantom{abc}& SSRN\cite{IEEE/journals/tgrs/Zhong2018}& A2S2K\cite{IEEE/journals/tgrs/Roy2020} & Triplet-Watershed\\
    \midrule
                   1  & 663  & 5968   &\phantom{abc}& 91.11 $\pm$ 0.007   & 94.30  $\pm$ 0.008  & 94.34 $\pm$ 0.0032 &\phantom{abc}& 99.85  $\pm$ 0.001           & 99.91  $\pm$ 0.000          & \textbf{100.0 $\pm$ 0.000}\\
    \rowcolor{Gray}2  & 1864 & 16785  &\phantom{abc}& 98.11 $\pm$ 0.003   & 97.65  $\pm$ 0.002  & 95.24 $\pm$ 0.0051 &\phantom{abc}& 99.98  $\pm$ 0.000           & 99.99  $\pm$ 0.000          & \textbf{100.0 $\pm$ 0.000}\\
                   3  & 209  & 1890   &\phantom{abc}& 67.71 $\pm$ 0.014   & 81.26  $\pm$ 0.018  & 69.39 $\pm$ 0.0151 &\phantom{abc}& 99.68  $\pm$ 0.003           & 99.88  $\pm$ 0.001          & \textbf{99.8 $\pm$ 0.004}\\
    \rowcolor{Gray}4  & 306  & 2758   &\phantom{abc}& 88.20 $\pm$ 0.006   & 94.63  $\pm$ 0.004  & 78.69 $\pm$ 0.0058 &\phantom{abc}& 99.92  $\pm$ 0.001           & 99.95  $\pm$ 0.001          & \textbf{99.96 $\pm$ 0.001}\\
                   5  & 134  & 1211   &\phantom{abc}& 98.93 $\pm$ 0.002   & 99.20  $\pm$ 0.002  & 87.46 $\pm$ 0.0110 &\phantom{abc}& 99.94  $\pm$ 0.000           & \textbf{100.0  $\pm$ 0.000} & \textbf{100.0 $\pm$ 0.000}\\
    \rowcolor{Gray}6  & 502  & 4527   &\phantom{abc}& 72.14 $\pm$ 0,022   & 90.58  $\pm$ 0,008  & 61.37 $\pm$ 0.0111 &\phantom{abc}& 99.95  $\pm$ 0.001           & 99.91  $\pm$ 0,001          & \textbf{99.99 $\pm$ 0.001}\\
                   7  & 133  & 1197   &\phantom{abc}& 75.69 $\pm$ 0.017   & 85.71  $\pm$ 0.011  & 75.49 $\pm$ 0.0295 &\phantom{abc}& \textbf{100.0  $\pm$ 0.000}  & \textbf{100.0  $\pm$ 0.000} & \textbf{100.0 $\pm$ 0.000}\\
    \rowcolor{Gray}8  & 368  & 3314   &\phantom{abc}& 89.64 $\pm$ 0.013   & 88.20  $\pm$ 0.003  & 74.65 $\pm$ 0.0044 &\phantom{abc}& 98.28  $\pm$ 0.015           & 98.88  $\pm$ 0.006          & \textbf{99.97 $\pm$ 0.001}\\
                   9  & 94   & 853    &\phantom{abc}& 99.77 $\pm$ 0.002   & 99.84  $\pm$ 0.001  & 99.77 $\pm$ 0.0015 &\phantom{abc}& 99.39  $\pm$ 0.003           & 99.78  $\pm$ 0.003          & \textbf{100.0 $\pm$ 0.000}\\
    \midrule
                   OA &4273  & 38503  &\phantom{abc}& 90.41 $\pm$ 0.001   & 94.19  $\pm$ 0.002  & 86.13 $\pm$ 0.0023 &\phantom{abc}& 99.77  $\pm$ 0.001  & 99.85  $\pm$ 0.001 & \textbf{99.98 $\pm$ 0.001}  \\
    \rowcolor{Gray}AA &      &        &\phantom{abc}& 86.81 $\pm$ 0.002   & 92.38  $\pm$ 0.003  & 81.82 $\pm$ 0.0039&\phantom{abc}& 99.66  $\pm$ 0.001  & 99.81  $\pm$ 0.001 &\textbf{99.97 $\pm$ 0.001} \\
    $\kappa$ &&        &\phantom{abc}& 0.8710 $\pm$ 0.002  & 0.9229 $\pm$ 0.002  & 0.8136 $\pm$ 0.0030 &\phantom{abc}& 0.9969 $\pm$ 0.001  & 0.9981 $\pm$ 0.001 & \textbf{0.9998 $\pm$ 0.001} \\    \bottomrule
    \end{tabular}
\end{adjustbox}
\end{table*}

\begin{table*}
\caption{Overall Accuracy (OA), Average Accuracy (AA), and Kappa($\kappa$) values on Kennedy Space Centre (KSC) dataset using 10\% of samples for training.}
\label{table:4}
\centering
\begin{adjustbox}{max width=.95\linewidth}
\begin{tabular}{ccccccccccc}

    \toprule
    \phantom{abc} & \phantom{abc} & \phantom{abc} &\phantom{abc}& \multicolumn{3}{c}{Classic approaches} & \phantom{abc} & \multicolumn{3}{c}{Deep-Learning approaches}\\
         \cmidrule{5-7}  \cmidrule{9-11}
    Class& Train & Test &\phantom{abc}& RF\cite{IEEE:journals/tgrs/Ham2005} & SVM\cite{DBLP:journals/tgrs/MelganiB04} & Ensemble-Watershed\cite{DBLP:journals/spl/ChallaDSN19} &\phantom{abc}& SSRN\cite{IEEE/journals/tgrs/Zhong2018}& A2S2K\cite{IEEE/journals/tgrs/Roy2020} & Triplet-Watershed\\
    \midrule
                   1 & 76 & 685 &\phantom{abc}& 94.79 $\pm$ 0.012 & 95.43 $\pm$ 0.023 &  96.23 $\pm$ 0.0085 &\phantom{abc}& 99.95 $\pm$ 0.001 & 99.95  $\pm$ 0.001                           &  \textbf{100.0 $\pm$ 0.0000}\\
    \rowcolor{Gray}2 & 24 & 219 &\phantom{abc}& 81.58 $\pm$ 0 047 & 83.71 $\pm$ 0.012 &  89.59 $\pm$ 0.0247 &\phantom{abc}& \textbf{100.0 $\pm$ 0.000} & 98.68  $\pm$ 0.019                  &  \textbf{100.0 $\pm$ 0.0000}\\
                   3 & 25 & 231 &\phantom{abc}& 86.09 $\pm$ 0 020 & 78.41 $\pm$ 0.218 &  83.98 $\pm$ 0.0341 &\phantom{abc}& 99.66 $\pm$ 0.005 & 98.72  $\pm$ 0.012                           &  \textbf{100.0 $\pm$ 0.0000}\\
    \rowcolor{Gray}4 & 25 & 227 &\phantom{abc}& 71.22 $\pm$ 0.061 & 27.17 $\pm$ 0.173 &  69.60 $\pm$ 0.0406 &\phantom{abc}& 91.22 $\pm$ 0.047 & 94.27  $\pm$ 0.042                           &  \textbf{96.56 $\pm$ 0.0423}\\
                   5 & 16 & 145 &\phantom{abc}& 47.59 $\pm$ 0.060 & 22.99 $\pm$ 0.170 &  65.52 $\pm$ 0.0474 &\phantom{abc}& \textbf{100.0 $\pm$ 0.000} & 94.46  $\pm$ 0.050                  &  99.86 $\pm$ 0.0028\\
    \rowcolor{Gray}6 & 22 & 207 &\phantom{abc}& 48.22 $\pm$ 0.014 & 36.89 $\pm$ 0.078 &  53.33 $\pm$ 0.0526 &\phantom{abc}& 98.45 $\pm$ 0.022 & 99.82  $\pm$ 0.003                           &  \textbf{99.52 $\pm$ 0.0000}\\
                   7 & 10 & 95  &\phantom{abc}& 79.43 $\pm$ 0 096 & 87.94 $\pm$ 0.027 &  85.05 $\pm$ 0.0234 &\phantom{abc}& 95.42 $\pm$ 0.050 & 99.61  $\pm$ 0.005                           &  \textbf{100.0 $\pm$ 0.0000}\\
    \rowcolor{Gray}8 & 43 & 388 &\phantom{abc}& 78.61 $\pm$ 0.054 & 70.19 $\pm$ 0.073 &  91.24 $\pm$ 0.0297 &\phantom{abc}& 99.80 $\pm$ 0.003 & \textbf{100.0  $\pm$ 0.000}                  &  \textbf{99.90 $\pm$ 0.0000}\\
                   9 & 52 & 468 &\phantom{abc}& 89.46 $\pm$ 0.011 & 85.33 $\pm$ 0.021 &  93.08 $\pm$ 0.0193 &\phantom{abc}& \textbf{100.0 $\pm$ 0.000} & \textbf{100.0  $\pm$ 0.000}         &  \textbf{100.0 $\pm$ 0.0000}\\
    \rowcolor{Gray}10 &40 & 364 &\phantom{abc}& 88.43 $\pm$ 0.034 & 78.88 $\pm$ 0.069 &  92.64 $\pm$ 0.0150 &\phantom{abc}& \textbf{100.0 $\pm$ 0.000} & \textbf{100.0  $\pm$ 0.000}         &  \textbf{100.0 $\pm$ 0.0000}\\
                   11 &41 & 378 &\phantom{abc}& 95.58 $\pm$ 0.014 & 93.81 $\pm$ 0.008 &  94.44 $\pm$ 0.0261 &\phantom{abc}& \textbf{100.0 $\pm$ 0.000} & \textbf{100.0  $\pm$ 0.000}         &  \textbf{100.0 $\pm$ 0.0000}\\
    \rowcolor{Gray}12 &50 & 453 &\phantom{abc}& 82.63 $\pm$ 0.032 & 86.98 $\pm$ 0.009 &  86.98 $\pm$ 0.0119 &\phantom{abc}& \textbf{100.0 $\pm$ 0.000} & \textbf{100.0  $\pm$ 0.000}         & 99.21 $\pm$ 0.0159\\
                   13 &92 & 835 &\phantom{abc}& 99.60 $\pm$ 0.002 & \textbf{100.0 $\pm$ 0.000} &  99.69 $\pm$ 0.0022 &\phantom{abc}& \textbf{100.0 $\pm$ 0.000} & \textbf{100.0  $\pm$ 0.000}&  \textbf{100.0 $\pm$ 0.0000} \\
    \midrule
                   OA & 516 & 4695  &\phantom{abc}& 86.17  $\pm$ 0.004  & 81.27 $\pm$ 0.008   & 89.54 $\pm$ 0.0038 &\phantom{abc}& 99.29 $\pm$ 0.004   & 99.34 $\pm$ 0.0008 & \textbf{99.72  $\pm$ 0.0023}\\
    \rowcolor{Gray}AA &      &      &\phantom{abc}& 80.25  $\pm$ 0.004  & 72.90 $\pm$ 0.021   & 84.72 $\pm$ 0.0038 &\phantom{abc}& 98.80 $\pm$ 0.008   & 98.88 $\pm$ 0.0018 & \textbf{99.62  $\pm$ 0.0032}\\
    $\kappa$ &&      &\phantom{abc}& 0.8459 $\pm$ 0.004  & 0.7909 $\pm$ 0.009  & 0.8834 $\pm$ 0.0042 &\phantom{abc}& 0.9921 $\pm$ 0.004  & 0.9927 $\pm$ 0.001 & \textbf{0.9969 $\pm$ 0.0026}\\
    \bottomrule
    \end{tabular}

\end{adjustbox}
\end{table*}

\begin{table*}

\caption{Overall Accuracy (OA), Average Accuracy (AA), and Kappa($\kappa$) values on University of Houston (UH) dataset using 10\% of samples for training.}
\label{table:R1_UH}
\centering
\begin{adjustbox}{max width=.95\linewidth}
\begin{tabular}{ccccccccccc}
    \toprule
        \phantom{abc} & \phantom{abc} & \phantom{abc} &\phantom{abc}& \multicolumn{3}{c}{Classic approaches} & \phantom{abc} & \multicolumn{3}{c}{Deep-Learning approaches}\\
         \cmidrule{5-7}  \cmidrule{9-11}
    Class& Train & Test &\phantom{abc}& RF\cite{IEEE:journals/tgrs/Ham2005} & SVM\cite{DBLP:journals/tgrs/MelganiB04} & Ensemble-Watershed\cite{DBLP:journals/spl/ChallaDSN19} &\phantom{abc}& SSRN\cite{IEEE/journals/tgrs/Zhong2018}& A2S2K\cite{IEEE/journals/tgrs/Roy2020} &Triplet-Watershed\\
    \midrule
                   1  & 125 & 1126 & \phantom{abc} & 82.52 $\pm$ 0.0000 & 82.33 $\pm$ 0.0000 &  93.68 $\pm$ 0.0279 & \phantom{abc} &99.66 $\pm$ 0.0012 & \textbf{99.79  $\pm$  0.0021} & 98.99 $\pm$ 0.0080 \\
    \rowcolor{Gray}2  & 125 & 1129 & \phantom{abc} & 83.30 $\pm$ 0.0011 & 83.36 $\pm$ 0.0000 &  81.97 $\pm$ 0.0191 & \phantom{abc} &99.96 $\pm$ 0.0004 & \textbf{100.0  $\pm$  0.0000} & \textbf{100.0 $\pm$ 0.0000} \\
                   3  & 69  & 628  & \phantom{abc} & 97.62 $\pm$ 0.0000 & 99.80 $\pm$ 0.0000 &  99.90 $\pm$ 0.0013 & \phantom{abc} &\textbf{100.0 $\pm$ 0.0000} & \textbf{100.0  $\pm$  0.0000} & \textbf{100.0 $\pm$ 00000} \\
    \rowcolor{Gray}4  & 124 & 1120 & \phantom{abc} & 91.41 $\pm$ 0.0027 & 98.95 $\pm$ 0.0000 &  74.27 $\pm$ 0.0240 & \phantom{abc} &99.66 $\pm$ 0.0046 & 99.17  $\pm$  0.0095 & \textbf{100.0 $\pm$ 0.0000} \\
                   5  & 124 & 1118 & \phantom{abc} & 96.49 $\pm$ 0.0020 & 98.76 $\pm$ 0.0000 &  82.15 $\pm$ 0.0214 & \phantom{abc} &\textbf{100.0 $\pm$ 0.0000} & \textbf{100.0  $\pm$  0.0000} & \textbf{100.0 $\pm$ 00000} \\
    \rowcolor{Gray}6  & 32  & 293  & \phantom{abc} & 99.30 $\pm$ 0.0000 & 97.90 $\pm$ 0.0000 &  92.22 $\pm$ 0.0613 & \phantom{abc} &\textbf{100.0 $\pm$ 0.0000} & \textbf{100.0  $\pm$  0.0000} & 99.43 $\pm$ 0.0080 \\
                   7  & 126 & 1142 & \phantom{abc} & 75.09 $\pm$ 0.0020 & 77.42 $\pm$ 0.0000 &  69.63 $\pm$ 0.0272 & \phantom{abc} &99.10 $\pm$ 0.0119 & 98.98  $\pm$  0.0088 & \textbf{99.65 $\pm$ 0.0050} \\
    \rowcolor{Gray}8  & 124 & 1120 & \phantom{abc} & 33.04 $\pm$ 0.0020 & 60.30 $\pm$ 0.0000 &  78.25 $\pm$ 0.0242 & \phantom{abc} &99.38 $\pm$ 0.0016 & \textbf{99.72  $\pm$  0.0038} & 96.25 $\pm$ 0.0338 \\
                   9  & 125 & 1127 & \phantom{abc} & 69.31 $\pm$ 0.0042 & 76.77 $\pm$ 0.0000 &  52.56 $\pm$ 0.0159 & \phantom{abc} &\textbf{99.30 $\pm$ 0.0052} & 98.47  $\pm$  0.0101 & 97.96 $\pm$ 0.0145 \\
    \rowcolor{Gray}10 & 122 & 1105 & \phantom{abc} & 44.11 $\pm$ 0.0034 & 61.29 $\pm$ 0.0000 &  63.66 $\pm$ 0.0207 & \phantom{abc} &94.85 $\pm$ 0.0152 & 94.90  $\pm$  0.0178 & \textbf{100.0 $\pm$ 0.0000} \\
                   11 & 123 & 1112 & \phantom{abc} & 70.20 $\pm$ 0.0020 & 80.55 $\pm$ 0.0000 &  56.83 $\pm$ 0.0379 & \phantom{abc} &99.23 $\pm$ 0.0075 & \textbf{99.42  $\pm$  0.0040} & 99.07 $\pm$ 0.0131 \\
    \rowcolor{Gray}12 & 123 & 1110 & \phantom{abc} & 54.81 $\pm$ 0.0036 & 79.92 $\pm$ 0.0000 &  54.77 $\pm$ 0.0319 & \phantom{abc} &98.76 $\pm$ 0.0028 & 99.46  $\pm$  0.0033 & \textbf{99.64 $\pm$ 0.0000} \\
                   13 & 46  & 423  & \phantom{abc} & 60.23 $\pm$ 0.0129 & 70.87 $\pm$ 0.0000 &  06.52 $\pm$ 0.0130 & \phantom{abc} &\textbf{99.90 $\pm$ 0.0013} & 99.01  $\pm$  0.0101 & 98.74 $\pm$ 0.0089 \\
    \rowcolor{Gray}14 & 42  & 386  & \phantom{abc} & 99.32 $\pm$ 0.0019 & 100.0 $\pm$ 0.0000 &  94.15 $\pm$ 0.0089 & \phantom{abc} &98.63 $\pm$ 0.0193 & \textbf{100.0  $\pm$  0.0000} & \textbf{100.0 $\pm$ 0.0000} \\
                   15 & 66  & 594  & \phantom{abc} & 97.25 $\pm$ 0.0017 & 96.40 $\pm$ 0.0000 &  98.55 $\pm$ 0.0051 & \phantom{abc} &\textbf{100.0 $\pm$ 0.0000} & \textbf{100.0  $\pm$  0.0000} & \textbf{100.0 $\pm$ 00000} \\
    \midrule               
    \rowcolor{Gray}OA &     &      & \phantom{abc} & 73.02 $\pm$ 0.0004 & 81.86 $\pm$ 0.0000 &  72.50 $\pm$ 0.0030 & \phantom{abc} &99.10 $\pm$ 0.0013 & 99.12  $\pm$  0.0030 & \textbf{99.25 $\pm$ 0.0039} \\
                   AA &     &      & \phantom{abc} & 76.93 $\pm$ 0.0004 & 84.31 $\pm$ 0.0000 &  73.27 $\pm$ 0.0046 & \phantom{abc} &99.23 $\pm$ 0.0016 & 99.26  $\pm$  0.0020 & \textbf{99.32 $\pm$ 0.0031} \\
    \rowcolor{Gray}$Kappa$& &      & \phantom{abc} & 71.01 $\pm$ 0.0003 & 80.42 $\pm$ 0.0000 &  70.22 $\pm$ 0.0033 & \phantom{abc} &99.03 $\pm$ 0.0015 & 99.05  $\pm$  0.0033 & \textbf{99.19 $\pm$ 0.0042} \\ 
       
    \bottomrule
    \end{tabular}

\end{adjustbox}
\end{table*}

\begin{table*}
\caption{Overall Accuracy (OA), Average Accuracy (AA), and Kappa($\kappa$) values on Indianpines (IP) dataset using Semi-Supervised approaches.}
\label{table:2Semi}
\centering
\begin{tabular}{ccccccccc}
\toprule
Class& Train & Test &\phantom{abc}& S2GCN\cite{IEEE:journal/grsl/Qin2019} & SSRN\cite{IEEE/journals/tgrs/Zhong2018} & DC-GCN\cite{ARXIV:Zeng2020} &  Triplet-Watershed\\
\midrule
                1  & 30 & 16   &\phantom{abc}&   \textbf{100.0 $\pm$ 0.0000} &   93.24 $\pm$ 0.0263 &  \textbf{100.00 $\pm$  0.0000} &  \textbf{100.00 $\pm$ 0.0000}  \\
\rowcolor{Gray} 2  & 30 & 1398 &\phantom{abc}&    84.43 $\pm$ 0.0250 &   76.63 $\pm$ 0.0596 &   91.28 $\pm$  0.0360 &   \textbf{91.69 $\pm$ 0.0194}  \\
                3  & 30 & 800  &\phantom{abc}&    82.87 $\pm$ 0.0553 &   68.78 $\pm$ 0.0753 &   92.88 $\pm$  0.0396 &  \textbf{95.25 $\pm$ 0.0610}  \\
\rowcolor{Gray} 4  & 30 & 207  &\phantom{abc}&    93.08 $\pm$ 0.0195 &   87.64 $\pm$ 0.0249 &   98.11 $\pm$  0.0151 &  \textbf{100.00 $\pm$ 0.0000}  \\
                5  & 30 & 453  &\phantom{abc}&    97.13 $\pm$ 0.0134 &   86.72 $\pm$ 0.0154 &   95.54 $\pm$  0.0339 &   \textbf{98.63 $\pm$ 0.0171}  \\
\rowcolor{Gray} 6  & 30 & 700  &\phantom{abc}&    97.29 $\pm$ 0.0127 &   92.05 $\pm$ 0.0182 &   98.67 $\pm$  0.0104 &  \textbf{100.00 $\pm$ 0.0000}  \\
                7  & 15 & 13   &\phantom{abc}&    92.31 $\pm$ 0.0000 &   95.66 $\pm$ 0.0051 &  \textbf{100.00 $\pm$  0.0000} &  \textbf{100.00 $\pm$ 0.0000}  \\
\rowcolor{Gray} 8  & 30 & 448  &\phantom{abc}&    99.03 $\pm$ 0.0093 &   95.90 $\pm$ 0.0297 &  \textbf{100.00 $\pm$  0.0000} &  \textbf{100.00 $\pm$ 0.0000}  \\
                9  & 15 & 5    &\phantom{abc}&   \textbf{100.00 $\pm$ 0.0000} &  \textbf{100.00 $\pm$ 0.0000} &  \textbf{100.00 $\pm$  0.0000} &  \textbf{100.00 $\pm$ 0.0000}  \\
\rowcolor{Gray} 10 & 30 & 942  &\phantom{abc}&    93.77 $\pm$ 0.0373 &   82.42 $\pm$ 0.0324 &   91.91 $\pm$  0.0378 &   \textbf{98.22 $\pm$ 0.0232}  \\
                11 & 30 & 2425 &\phantom{abc}&    84.98 $\pm$ 0.0282 &   82.23 $\pm$ 0.0288 &   91.79 $\pm$  0.0379 &   \textbf{94.43 $\pm$ 0.0229}  \\
\rowcolor{Gray} 12 & 30 & 563  &\phantom{abc}&    80.05 $\pm$ 0.0517 &   69.09 $\pm$ 0.0436 &   90.17 $\pm$  0.0554 &   \textbf{99.08 $\pm$ 0.0185}  \\
                13 & 30 & 175  &\phantom{abc}&    99.43 $\pm$ 0.0000 &   95.78 $\pm$ 0.0075 &   99.65 $\pm$  0.0027 &  \textbf{100.00 $\pm$ 0.0000}  \\
\rowcolor{Gray} 14 & 30 & 1235 &\phantom{abc}&    96.73 $\pm$ 0.0092 &   86.52 $\pm$ 0.0243 &   99.73 $\pm$  0.0066 &  \textbf{99.87 $\pm$ 0.0026}  \\
                15 & 30 & 356  &\phantom{abc}&    86.80 $\pm$ 0.0342 &   73.12 $\pm$ 0.0528 &   99.94 $\pm$  0.0016 &  \textbf{100.00 $\pm$ 0.0000}  \\
\rowcolor{Gray} 16 & 30 & 63   &\phantom{abc}&   \textbf{100.00 $\pm$ 0.0000} &   86.21 $\pm$ 0.0130 &  \textbf{100.00 $\pm$  0.0000} &   99.37 $\pm$ 0.0078  \\
\midrule
                OA &      &      &\phantom{abc}&  89.4 $\pm$ 0.0108  &  88.34 $\pm$ 0.0173 &   94.65 $\pm$  0.1210 &   \textbf{ 96.74 $\pm$ 0.0194}   \\
\rowcolor{Gray} AA &      &      &\phantom{abc}&  92.9 $\pm$ 0.0104  &  85.75 $\pm$ 0.0069 &   96.85 $\pm$  0.0040 &   \textbf{98.53 $\pm$ 0.0098}   \\
          $\kappa$ &      &      &\phantom{abc}& 0.880 $\pm$ 0.012   & 0.866  $\pm$ 0.019  & 0.944   $\pm$  0.014  &  \textbf{0.9627 $\pm$ 0.0221}  \\
\bottomrule
\end{tabular}
\end{table*}

\begin{table*}
  \renewcommand{\arraystretch}{1.1}
\caption{Overall Accuracy (OA), Average Accuracy (AA), and Kappa($\kappa$) values on University of Pavia (UP) dataset using Semi-Supervised approaches.}
\label{table:3Semi}
\centering
\begin{tabular}{ccccccccc|}
\toprule
Class& Train & Test &\phantom{abc}& S2GCN\cite{IEEE:journal/grsl/Qin2019} & SSRN\cite{IEEE/journals/tgrs/Zhong2018} & DC-GCN\cite{ARXIV:Zeng2020} &  Triplet-Watershed\\
\midrule
                1  & 30  & 6601    &\phantom{abc}&   92.78 $\pm$ 0.0379   &  98.80 $\pm$ 0.0110  &  92.85 $\pm$ 0.0351   &   \textbf{99.56   $\pm$ 0.0088}  \\
\rowcolor{Gray} 2  & 30  & 18619   &\phantom{abc}&   87.06 $\pm$ 0.0447   &  98.45 $\pm$ 0.0054  &  97.53 $\pm$ 0.0140   &  \textbf{100.00   $\pm$ 0.0000}  \\
                3  & 30  & 2069    &\phantom{abc}&   87.97 $\pm$ 0.0477   &  77.05 $\pm$ 0.1024  &  97.94 $\pm$ 0.0118   &   \textbf{99.85   $\pm$ 0.0084}  \\
\rowcolor{Gray} 4  & 30  & 3034    &\phantom{abc}&   90.85 $\pm$ 0.0094   &  83.02 $\pm$ 0.0907  &  94.57 $\pm$ 0.0109   &   \textbf{99.99   $\pm$ 0.0003}  \\
                5  & 30  & 1315    &\phantom{abc}&  \textbf{100.00 $\pm$ 0.0000}   &  99.96 $\pm$ 0.0009  &  99.49 $\pm$ 0.0068   &  \textbf{100.00   $\pm$ 0.0000}  \\
\rowcolor{Gray} 6  & 30  & 4999    &\phantom{abc}&   88.69 $\pm$ 0.0264   &  87.03 $\pm$ 0.0626  &  98.57 $\pm$ 0.0278   &   \textbf{99.99   $\pm$ 0.0001}  \\
                7  & 30  & 1300    &\phantom{abc}&   98.88 $\pm$ 0.0108   &  83.92 $\pm$ 0.0897  & \textbf{100.00 $\pm$ 0.0000}   &  \textbf{100.00   $\pm$ 0.0000}  \\
\rowcolor{Gray} 8  & 30  & 3652    &\phantom{abc}&   89.97 $\pm$ 0.0328   &  88.41 $\pm$ 0.0463  &  \textbf{96.00 $\pm$ 0.0277}   &   92.15 $\pm$ 0.1560  \\
                9  & 30  & 917     &\phantom{abc}&   98.89 $\pm$ 0.0053   &  99.97 $\pm$ 0.0004  &  97.51 $\pm$ 0.0140   &  \textbf{100.00   $\pm$ 0.0000}  \\
\midrule
                OA &      &        &\phantom{abc}&   89.74 $\pm$ 0.0170   &  92.81 $\pm$ 0.0190  &  96.87 $\pm$ 0.0111   &   \textbf{99.20 $\pm$ 0.0129} \\
\rowcolor{Gray} AA &      &        &\phantom{abc}&   92.80 $\pm$ 0.0047   &  90.73 $\pm$ 0.0226  &  97.16 $\pm$ 0.0076   &   \textbf{98.95 $\pm$ 0.0165} \\
                $\kappa$ &&        &\phantom{abc}&  0.8665 $\pm$ 0.020    & 0.9059 $\pm$ 0.024   & 0.9677 $\pm$ 0.012    &   \textbf{0.9894 $\pm$ 0.0170} \\
\bottomrule
\end{tabular}
\end{table*}

In all the experiments we use the neural net architecture as shown in figure \ref{fig:3a}. We consider a patch ($11\times 11 \times \text{\#Bands}$) around each pixel of the input hyperspectral image, suitably padded with $0$s.  We use $3$ conv2d layers and a fully-connected layer to obtain the representation. These representations are then used for watershed classification and training. All models are trained using stochastic gradient descent (SGD) with cyclic learning rates\cite{DBLP:conf/wacv/Smith17}. We use $40\%$ of the training data as seeds for the watershed classifier. The default weight initialization by pytorch \cite{GIT:conference/nips/pytorch} is used. We use $64$ as the dimension for the representations. All accuracies are reported in the format $\text{mean}\times 100 \% \pm \text{stdev}$ to be consistent with \cite{IEEE/journals/tgrs/Roy2020}. The code is available at \url{https://github.com/ac20/TripletWatershed_Code}.

{\noindent

\textbf{Remark on evaluation:} Different kind of evaluations of possible - Random train/test split or Patch-based evaluation as proposed in \cite{IEEE:journals/grsl/Nalepa2019}. Here we use the former since - (i) Patch-based evaluation does not recommend using connectivity patterns, while watershed classifier is designed to exploit such patterns, (ii) Irrespective of the evaluation procedure, we remain consistent with baseline methods (A2S2K, SSRN). Hence, the observations in this article still remain valid. 
}

\begin{figure*}[!t]
\centering
\subfloat[IP]{\includegraphics[width=0.4\linewidth]{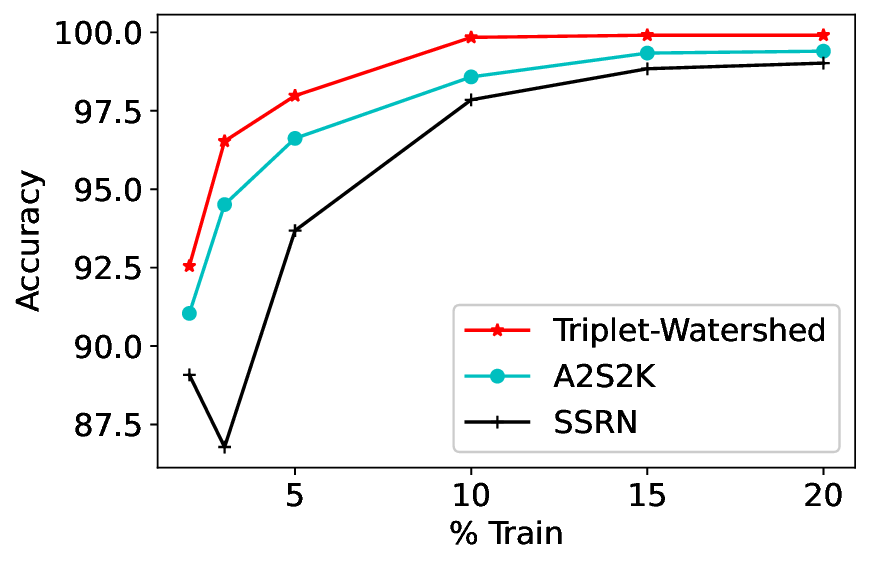}%
\label{fig:5a}}
\hfil
\subfloat[UP]{\includegraphics[width=0.4\linewidth]{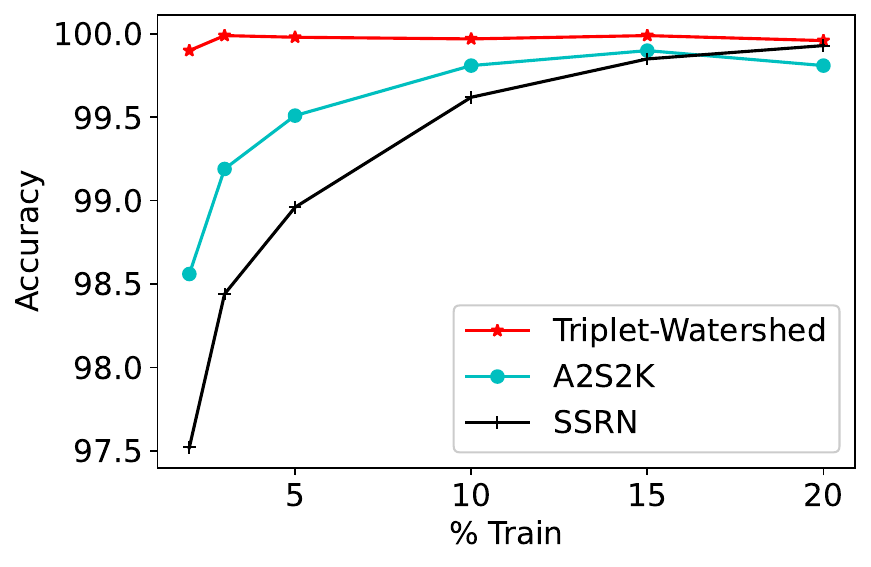}%
\label{fig:5b}}

\subfloat[KSC]{\includegraphics[width=0.4\linewidth]{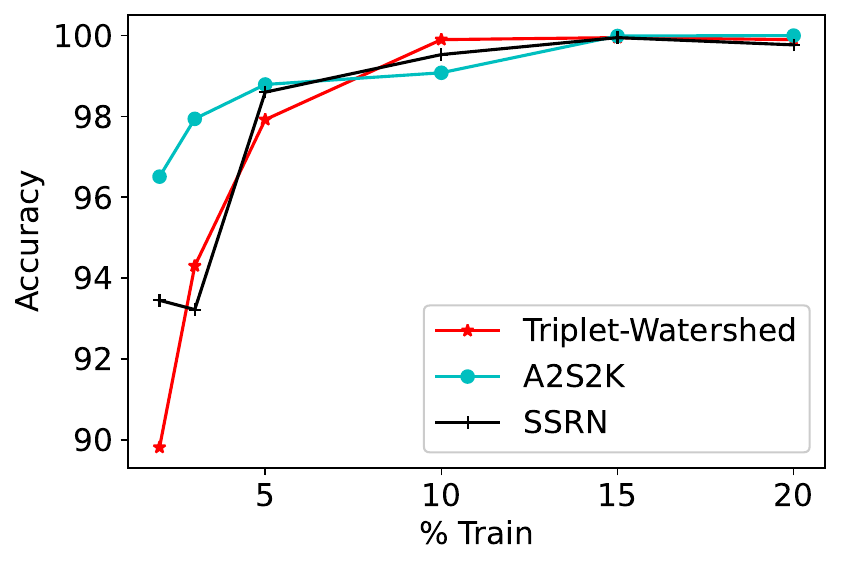}%
\label{fig:5c}}
\hfil
\subfloat[UH]{\includegraphics[width=0.4\linewidth]{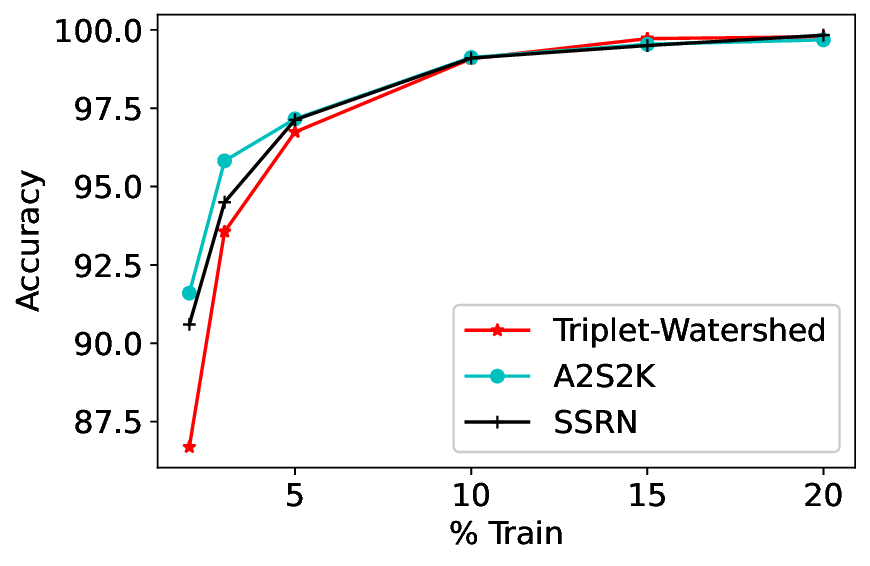}%
\label{fig:5d}}
\caption{Overall Accuracy (OA) vs \% training data. We observe that Triplet-Watershed outperforms other approaches even at small sizes of training data for Indianpines and University of Pavia Dataset. { IP denotes Indianpines dataset, UP denotes University of Pavia dataset, KSC denotes Kennedy Space Centre dataset and UH denotes University of Houston dataset}.}
\label{fig:5}
\end{figure*}

\begin{table}
  \renewcommand{\arraystretch}{1.1}
\caption{Comparison of Triplet-Watershed with Triplet-Random-Forest and Triplet-K-Nearest-Neighbors. {Replace Watershed classifier with Random Forest and KNN classifier to understand the importance of watershed classifier}. }
\label{table:5}
\centering
\begin{tabular}{cccc}
    \toprule
     & Triplet-Watershed & Triplet-RF & Triplet-KNN \\
    \midrule
    IN & \textbf{99.57 $\pm$ 0.0026} & 91.46 $\pm$ 0.011 & 90.86 $\pm$ 0.013 \\
    \rowcolor{Gray}UP & \textbf{99.98 $\pm$ 0.001} & 98.06 $\pm$ 0.007 & 99.62 $\pm$ 0.000 \\
    KSC& \textbf{99.72 $\pm$ 0.0023} & 87.80 $\pm$ 0.039 & 82.38 $\pm$ 0.031 \\
    \rowcolor{Gray}UH & {\textbf{99.25 $\pm$ 0.004}} & {89.02 $\pm$ 0.018} &  {96.15 $\pm$ 0.0086}\\
    \bottomrule
    \end{tabular}
\end{table}

\begin{table}
  \renewcommand{\arraystretch}{1.1}
\caption{Mean Average Precision (MAP) scores for the representations. { Observe that Triplet-Watershed obtains better representations than competing approaches on all datasets}.}
\label{table:6}
\centering
\begin{tabular}{cccc}
    \toprule
     & Triplet-Watershed & A2S2K\cite{IEEE/journals/tgrs/Roy2020} & SSRN\cite{IEEE/journals/tgrs/Zhong2018} \\
    \midrule
    IN & \textbf{0.9819} & 0.9713 & 0.9135 \\
    \rowcolor{Gray}UP &\textbf{0.9970} & 0.9821 & 0.9703 \\
    KSC& \textbf{0.9822} & 0.9837 & 0.9846 \\
    \rowcolor{Gray}UH& {\textbf{0.9821}} & {0.9799} & {0.9692} \\
    \bottomrule
\end{tabular}

% \begin{tabular}{cccc}
%     \toprule
%      & Triplet-Watershed & A2S2K\cite{IEEE/journals/tgrs/Roy2020} & SSRN\cite{IEEE/journals/tgrs/Zhong2018} \\
%     \midrule
%     IN & \textbf{0.9819} & 0.9713 & 0.9135 \\
%     \rowcolor{Gray}UP &\textbf{0.9970} & 0.9821 & 0.9703 \\
%     KSC& \textbf{0.9822} & 0.9837 & 0.9846 \\
%     \bottomrule
% \end{tabular}

% \begin{tabular}{cccc}
%     \toprule
%      & Triplet-Watershed & A2S2K\cite{IEEE/journals/tgrs/Roy2020} & SSRN\cite{IEEE/journals/tgrs/Zhong2018} \\
%     \midrule
%     IN & \textbf{0.9778} & 0.9713 & 0.9135 \\
%     \rowcolor{Gray}UP &\textbf{0.9911} & 0.9821 & 0.9703 \\
%     KSC& \textbf{0.9876} & 0.9837 & 0.9846 \\
%     \bottomrule
% \end{tabular}
\end{table}

\begin{table}
  \renewcommand{\arraystretch}{1.1}
\caption{Triplet-Watershed: Accuracy vs Embed Dimension. { Note that differences across various embedding dimensions are not significant}.}
\label{table:7}
\centering
\begin{adjustbox}{max width=.95\columnwidth}
\begin{tabular}{c c c c c c}
    \toprule
    Dimension & \phantom{abc} & KSC &IN & UP & UH \\
    \midrule
                   16 & \phantom{abc} & 99.53 $\pm$ 0.0031& 99.45 $\pm$ 0.0025 & 99.95 $\pm$ 0.0002 & {98.74 $\pm$ 0.0034}\\
    \rowcolor{Gray}32 & \phantom{abc} & 99.70 $\pm$ 0.0029& 99.72 $\pm$ 0.0010 & 99.97 $\pm$ 0.0003 & {98.73 $\pm$ 0.0018} \\
                   64 & \phantom{abc} & 99.54 $\pm$ 0.0017& 99.67 $\pm$ 0.0011 & 99.98 $\pm$ 0.0001 & {99.25 $\pm$ 0.0039}\\
    \rowcolor{Gray}128& \phantom{abc} & 99.72 $\pm$ 0.0004& 99.84 $\pm$ 0.0009 & 99.97 $\pm$ 0.0001 & {98.87 $\pm$ 0.0025}\\
    \bottomrule
    \end{tabular}

% \begin{tabular}{c c c c c}
% \toprule
% Dimension & \phantom{abc} & KSC &IN & UP \\
% \midrule
%                16 & \phantom{abc} & 99.53 $\pm$ 0.0031& 99.45 $\pm$ 0.0025 & 99.95 $\pm$ 0.0002\\
% \rowcolor{Gray}32 & \phantom{abc} & 99.70 $\pm$ 0.0029& 99.72 $\pm$ 0.0010 & 99.97 $\pm$ 0.0003\\
%                64 & \phantom{abc} & 99.54 $\pm$ 0.0017& 99.67 $\pm$ 0.0011 & 99.98 $\pm$ 0.0001\\
% \rowcolor{Gray}128& \phantom{abc} & 99.72 $\pm$ 0.0004& 99.84 $\pm$ 0.0009 & 99.97 $\pm$ 0.0001\\
% \bottomrule
% \end{tabular}

\end{adjustbox}
\end{table}

\begin{table}
  
  \renewcommand{\arraystretch}{1.1}
\caption{Run-times (in seconds) of Triplet-Watershed and other approaches. Observe that the running time of Triplet-Watershed is comparable to other approaches.}
\label{table:R1_8}
\centering
\begin{adjustbox}{max width=.95\columnwidth}
\begin{tabular}{ccccc}
    \toprule
     &Time(s) & Triplet-Watershed & A2S2K\cite{IEEE/journals/tgrs/Roy2020} & SSRN\cite{IEEE/journals/tgrs/Zhong2018} \\
    \midrule
    IN & Train &  520.56 & 829.23 & 779.33 \\
    \rowcolor{Gray}& Test &  3.77 & 10.55 & 11.44 \\
    UP & Train &791.22 & 2582.31 & 1964.66 \\
    \rowcolor{Gray}& Test &46.23 & 47.33 & 33.02 \\
    KSC& Train & 978.25 & 757.46 & 535.20 \\
    \rowcolor{Gray}& Test & 1.58 & 8.37 & 5.84 \\
    UH & Train & 1460.15 & 947.73 & 1145.38 \\
    \rowcolor{Gray}& Test & 8.74 & 11.55 & 7.85 \\
    \bottomrule
\end{tabular}
\end{adjustbox}
\end{table}

\begin{table}
  
\renewcommand{\arraystretch}{1.1}
\caption{Triplet-Watershed: Accuracy vs Patch Size.}
\label{table:R2_9}
\centering
\begin{adjustbox}{max width=.95\columnwidth}
\begin{tabular}{cccccc}
    \toprule
        & &7 & 9 & 11 & 13\\
    \midrule
    IN                & OA       & 99.72 $\pm$ 0.0021 & 99.56 $\pm$ 0.0021 & 99.63 $\pm$ 0.0017 & 99.63 $\pm$ 0.0022 \\
    \rowcolor{Gray}   & AA       & 99.72 $\pm$ 0.0024 & 98.57 $\pm$ 0.0180 & 99.82 $\pm$ 0.0012 & 99.75 $\pm$ 0.0009 \\
                      & $\kappa$ & 0.9968 $\pm$ 0.0024 & 0.9949 $\pm$ 0.0023 & 0.9958 $\pm$ 0.0020 & 0.9957 $\pm$ 0.0025 \\
    \rowcolor{Gray}UP & OA       & 99.96 $\pm$ 0.0008 & 99.98 $\pm$ 0.0002 & 99.99 $\pm$ 0.0002 & 99.98 $\pm$ 0.0002 \\
                      & AA       & 99.93 $\pm$ 0.0012 & 99.96 $\pm$ 0.0005 & 99.98 $\pm$ 0.0004 & 99.96 $\pm$ 0.0005 \\
    \rowcolor{Gray}   & $\kappa$ & 0.9994 $\pm$ 0.0010 & 0.9997 $\pm$ 0.0003 & 0.9999 $\pm$ 0.0002 & 0.9997 $\pm$ 0.0003 \\
    KSC               & OA       & 99.77 $\pm$ 0.0023 & 99.96 $\pm$ 0.0010 & 99.96 $\pm$ 0.0010 & 99.96 $\pm$ 0.0010 \\
    \rowcolor{Gray}   & AA       & 99.55 $\pm$ 0.0050 & 99.95 $\pm$ 0.0010 & 99.95 $\pm$ 0.0010 & 99.95 $\pm$ 0.0010 \\
                      & $\kappa$ & 0.9975 $\pm$ 0.0025 & 0.9995 $\pm$ 0.0010 & 0.9995 $\pm$ 0.0010 & 0.9995 $\pm$ 0.0010 \\
    \rowcolor{Gray}UH & OA       & 98.22 $\pm$ 0.0024 & 98.78 $\pm$ 0.0014 & 99.25 $\pm$ 0.0011 & 99.23 $\pm$ 0.0031 \\
                      & AA       & 98.28 $\pm$ 0.0038 & 98.89 $\pm$ 0.0015 & 99.32 $\pm$ 0.0013 & 99.26 $\pm$ 0.0031 \\
    \rowcolor{Gray}   & $\kappa$ & 0.9807 $\pm$ 0.0026 & 0.9868 $\pm$ 0.0015 & 0.9919 $\pm$ 0.0012 & 0.9915 $\pm$ 0.0034 \\
    \bottomrule
    \end{tabular}

\end{adjustbox}
\end{table}

\begin{figure*}[!t]
  \tiny
  \centering
  \subfloat[Triplet-Watershed-IP]{\includegraphics[width=0.24\linewidth]{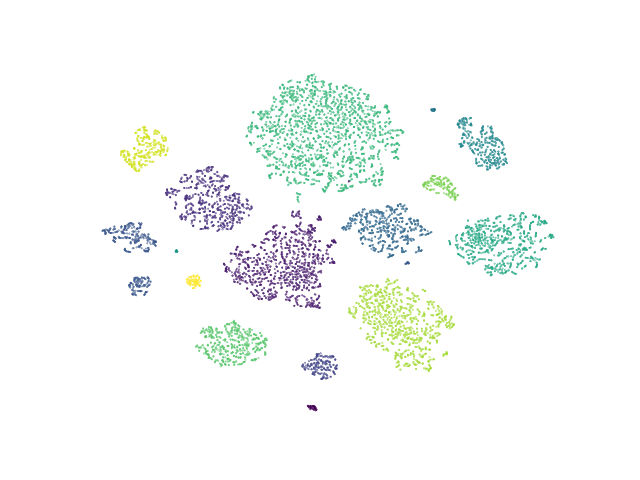}%
  \label{fig:6a}}
  \hfil
  \subfloat[Triplet-Watershed-UP]{\includegraphics[width=0.24\linewidth]{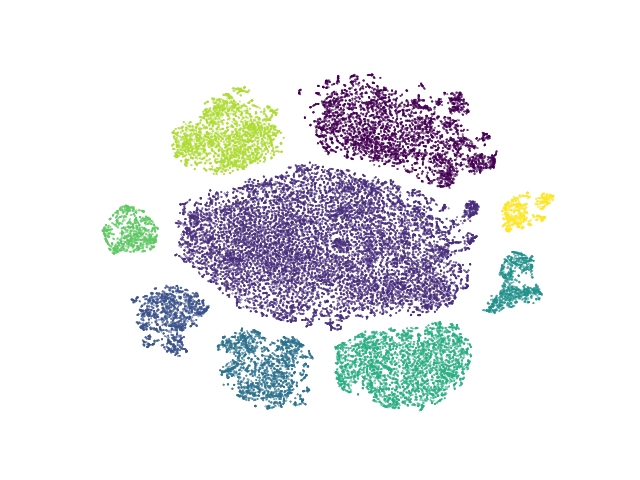}%
  \label{fig:6b}}
  \hfil
  \subfloat[Triplet-Watershed-KSC]{\includegraphics[width=0.24\linewidth]{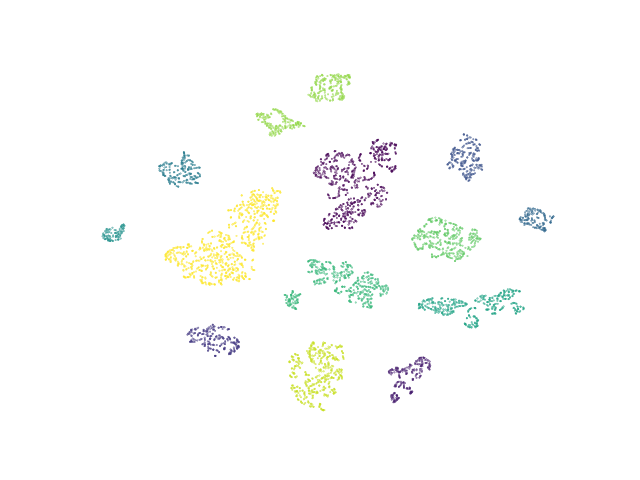}%
  \label{fig:6c}}
  \hfil
  \subfloat[Triplet-Watershed-UH]{\includegraphics[width=0.24\linewidth]{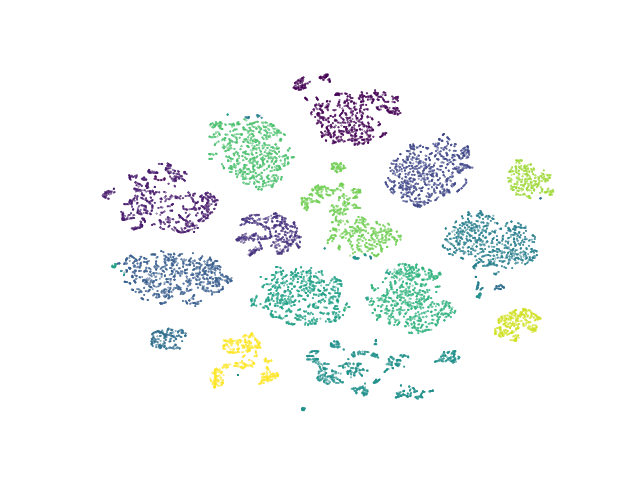}%
  \label{fig:R1_6a}}

  \subfloat[A2S2K-IP]{\includegraphics[width=0.24\linewidth]{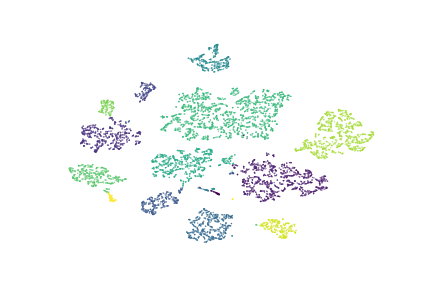}%
  \label{fig:6d}}
  \hfil
  \subfloat[A2S2K-UP]{\includegraphics[width=0.24\linewidth]{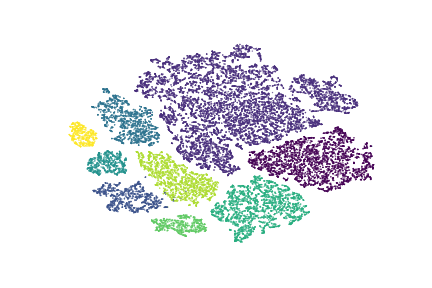}%
  \label{fig:6e}}
  \hfil
  \subfloat[A2S2K-KSC]{\includegraphics[width=0.24\linewidth]{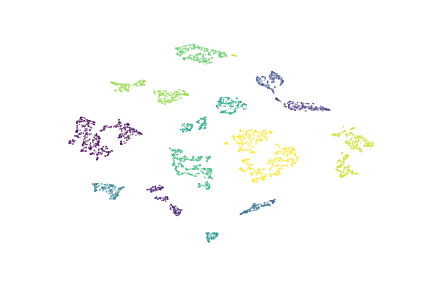}%
  \label{fig:6f}}
  \hfil
  \subfloat[A2S2K-UH]{\includegraphics[width=0.24\linewidth]{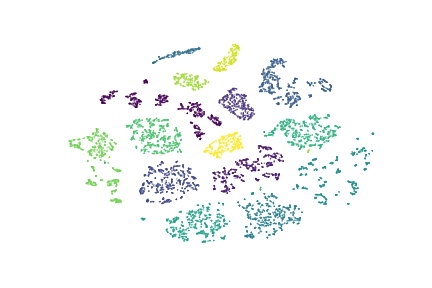}%
  \label{fig:R1_6d}}

  \subfloat[SSRN-IP]{\includegraphics[width=0.24\linewidth]{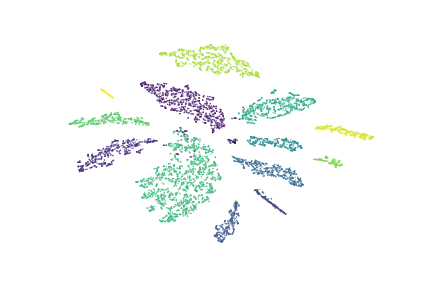}%
  \label{fig:6g}}
  \hfil
  \subfloat[SSRN-UP]{\includegraphics[width=0.24\linewidth]{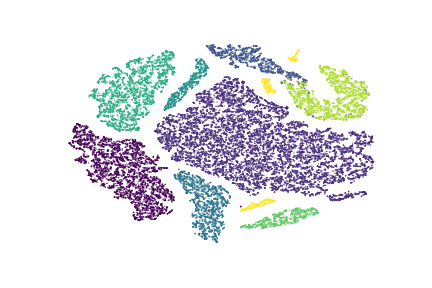}%
  \label{fig:6h}}
  \hfil
  \subfloat[SSRN-KSC]{\includegraphics[width=0.24\linewidth]{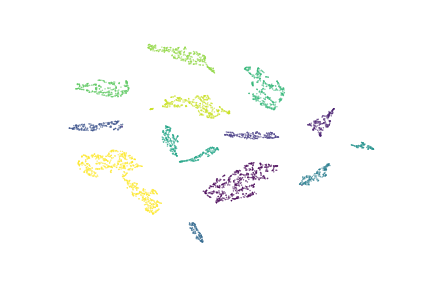}%
  \label{fig:6i}}
  \hfil
  \subfloat[SSRN-UH]{\includegraphics[width=0.24\linewidth]{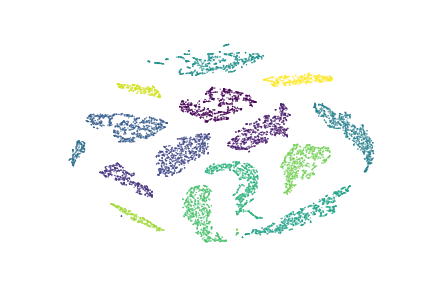}%
  \label{fig:R1_6g}}

  \caption{T-SNE Scatterplot of the various representations obtained. All approaches provide well-separated clusters, relatively compact. Table \ref{table:6} however shows that triplet-watershed achieves a better precision (MAP score). { IP denotes Indianpines dataset, UP denotes University of Pavia dataset, KSC denotes Kennedy Space Centre dataset and UH denotes University of Houston dataset}.}
  \label{fig:6}
\end{figure*}

\begin{figure*}
  \tiny
  \centering
  % \subfloat[GT]{\includegraphics[width=0.24\linewidth]{./img/FigureR1_7/indianpines_gt.png}%
  % \label{fig:R1_7_indianpinesa}}
  % \hfil
  % \subfloat[Triplet-Watershed]{\includegraphics[width=0.24\linewidth]{./img/FigureR1_7/indianpines_TripletWatershed.png}%
  % \label{fig:R1_7_indianpinesb}}
  % \hfil
  % \subfloat[SSRN]{\includegraphics[width=0.24\linewidth]{./img/FigureR1_7/indianpines_SSRN.png}%
  % \label{fig:R1_7_indianpinesc}}
  % \hfil
  % \subfloat[A2S2KResNet]{\includegraphics[width=0.24\linewidth]{./img/FigureR1_7/indianpines_A2S2K.png}%
  % \label{fig:R1_7_indianpinesd}}
  \subfloat[GT]{\includegraphics[width=0.24\linewidth]{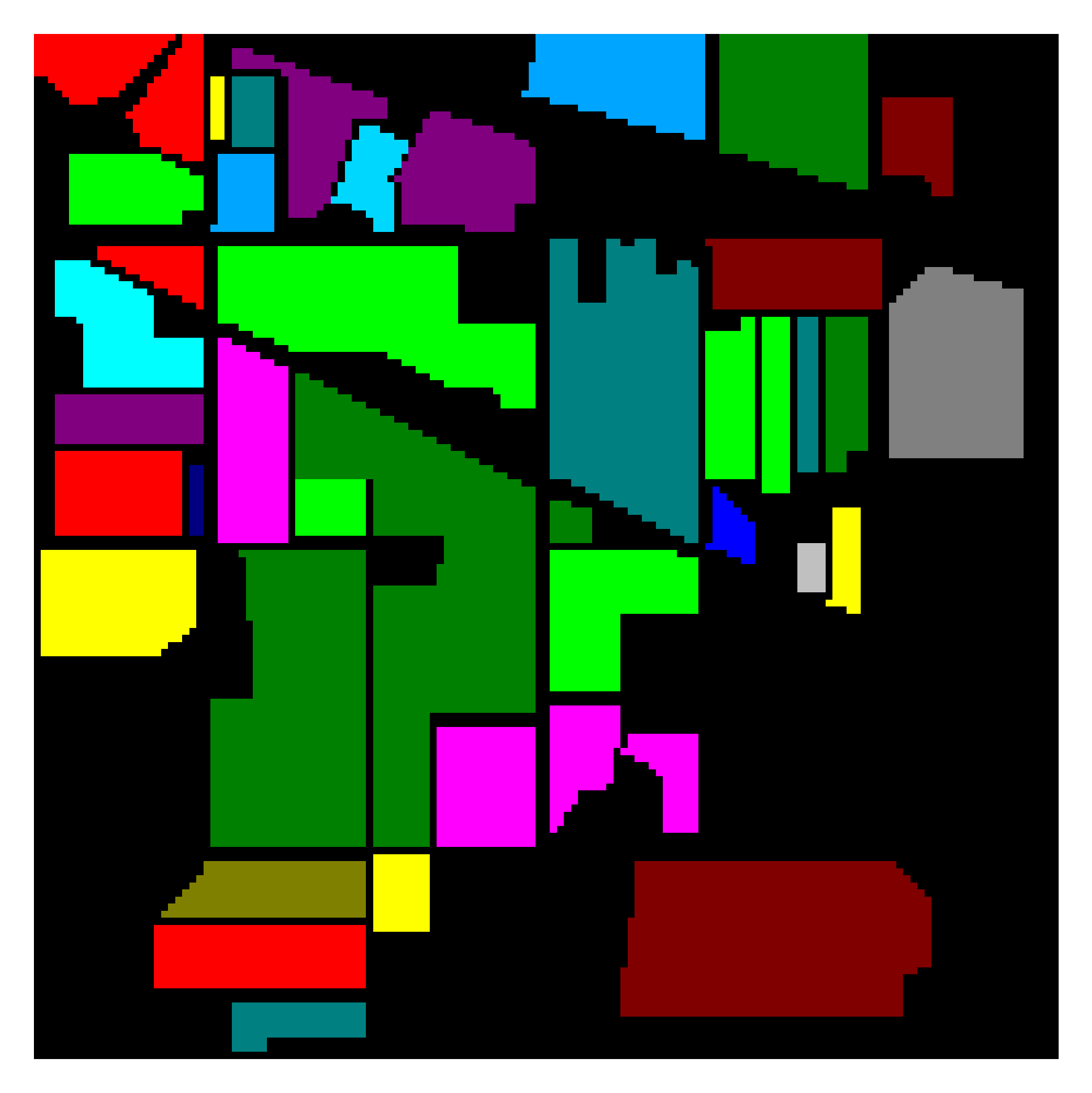}%
  \label{fig:R1_7_indianpinesa}}
  \hfil
  \subfloat[Triplet-Watershed]{\includegraphics[width=0.24\linewidth]{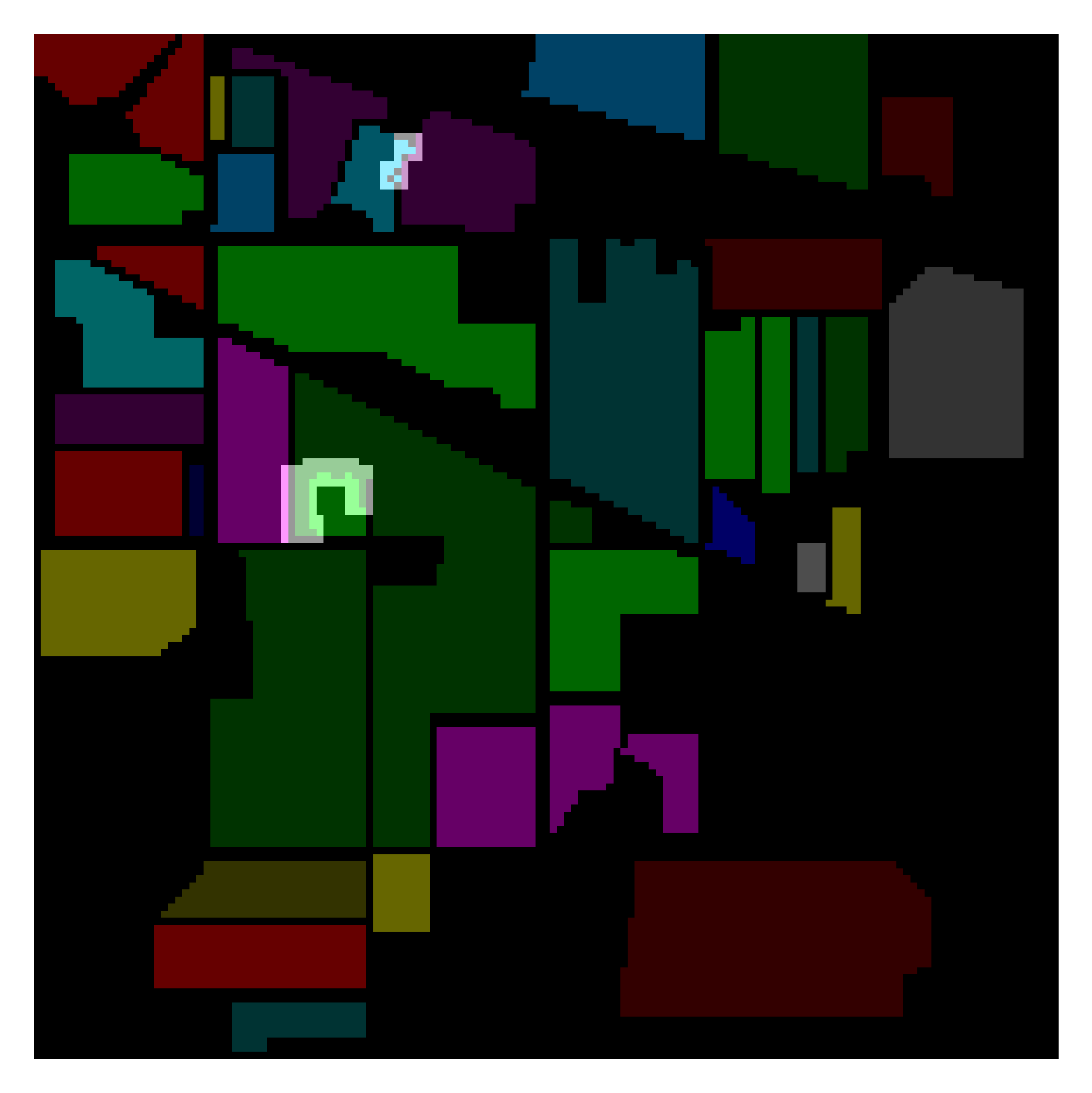}%
  \label{fig:R1_7_indianpinesb}}
  \hfil
  \subfloat[SSRN]{\includegraphics[width=0.24\linewidth]{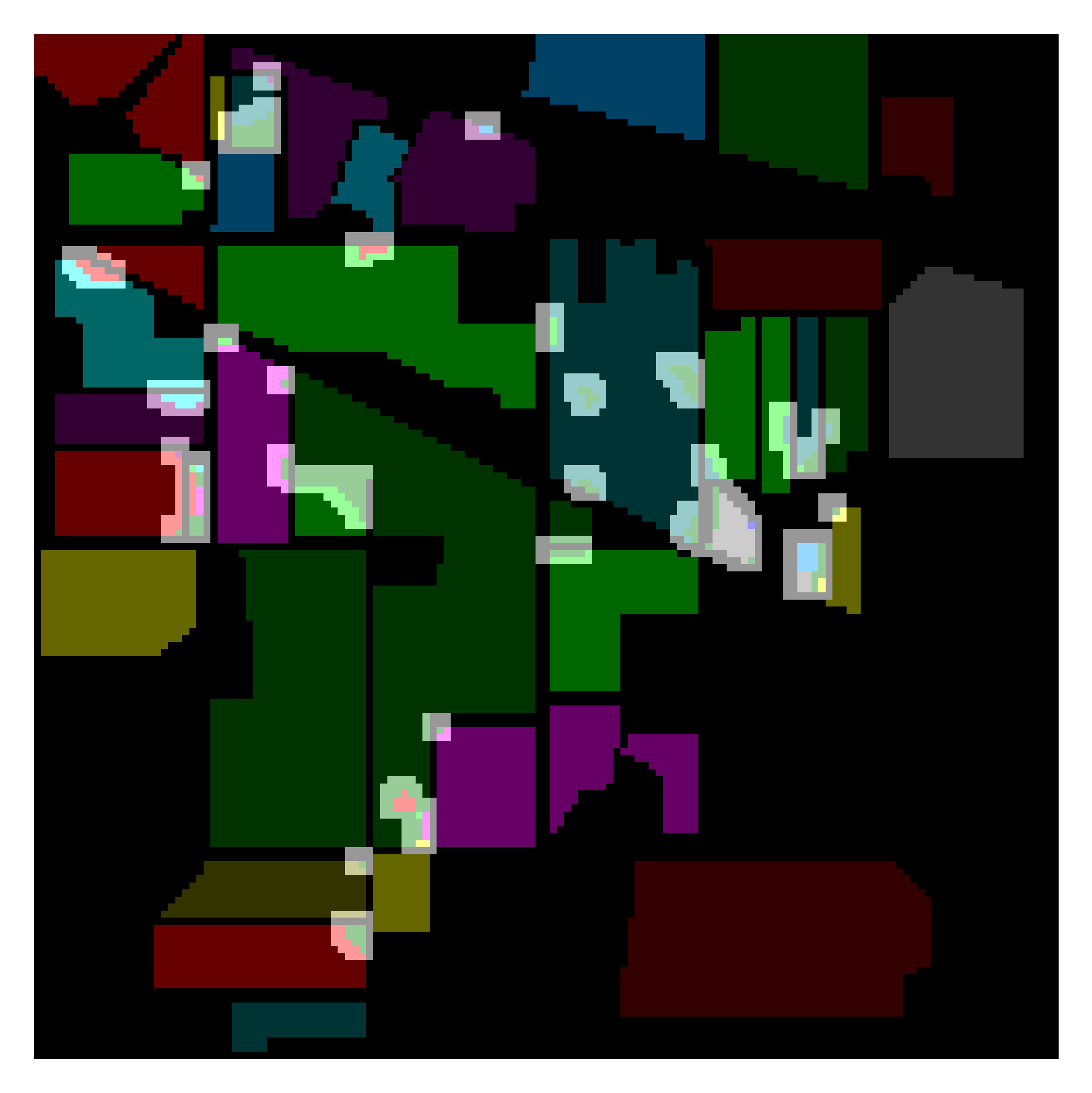}%
  \label{fig:R1_7_indianpinesc}}
  \hfil
  \subfloat[A2S2KResNet]{\includegraphics[width=0.24\linewidth]{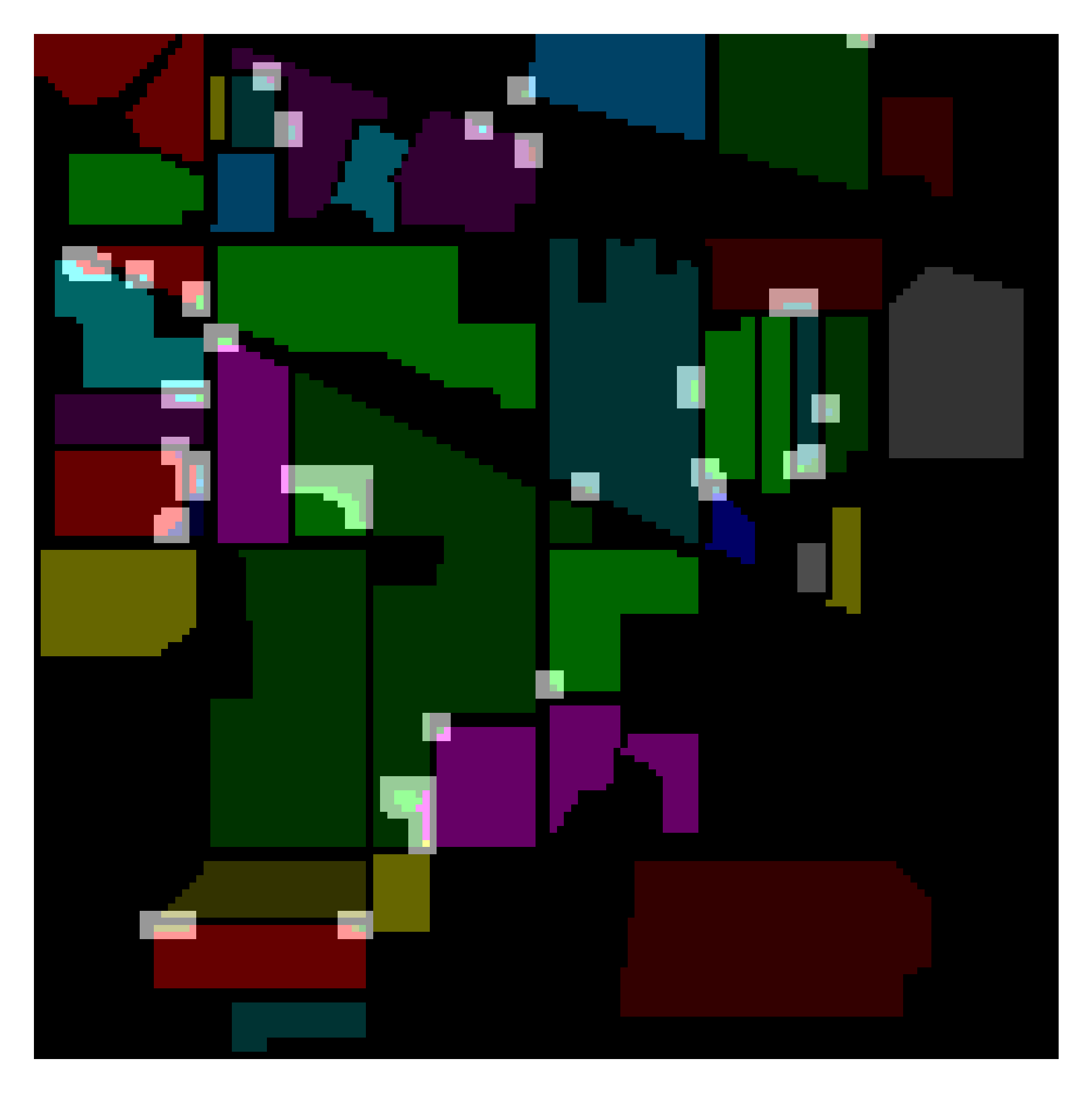}%
  \label{fig:R1_7_indianpinesd}}
  \caption{Classification maps for Indianpines (IP) dataset. The main differences with respect to groundtruth have been highlighted. As one can observe, the number of errors of Triplet-Watershed is small compared to SSRN and A2S2K.}
  \label{fig:R1_7_indianpines}
\end{figure*}

\begin{figure*}
  \tiny
  \centering
  % \subfloat[GT]{\includegraphics[width=0.24\linewidth]{./img/FigureR1_7/ksc_gt.png}%
  % \label{fig:R1_7_ksca}}
  % \hfil
  % \subfloat[Triplet-Watershed]{\includegraphics[width=0.24\linewidth]{./img/FigureR1_7/ksc_TripletWatershed.png}%
  % \label{fig:R1_7_kscb}}
  % \hfil
  % \subfloat[SSRN]{\includegraphics[width=0.24\linewidth]{./img/FigureR1_7/ksc_SSRN.png}%
  % \label{fig:R1_7_kscc}}
  % \hfil
  % \subfloat[A2S2KResNet]{\includegraphics[width=0.24\linewidth]{./img/FigureR1_7/ksc_A2S2K.png}%
  % \label{fig:R1_7_kscd}}
  \subfloat[GT]{\includegraphics[width=0.24\linewidth]{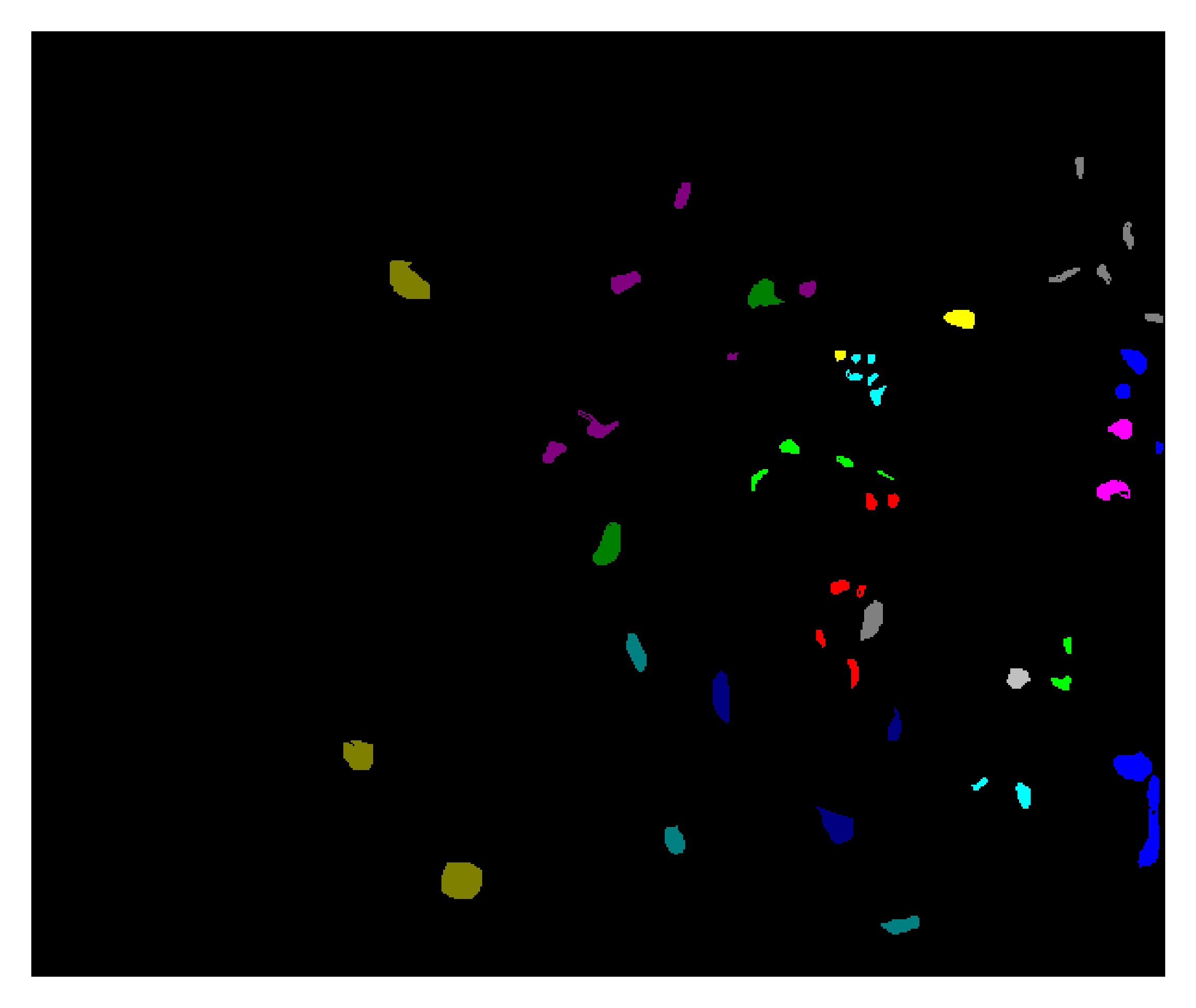}%
  \label{fig:R1_7_ksca}}
  \hfil
  \subfloat[Triplet-Watershed]{\includegraphics[width=0.24\linewidth]{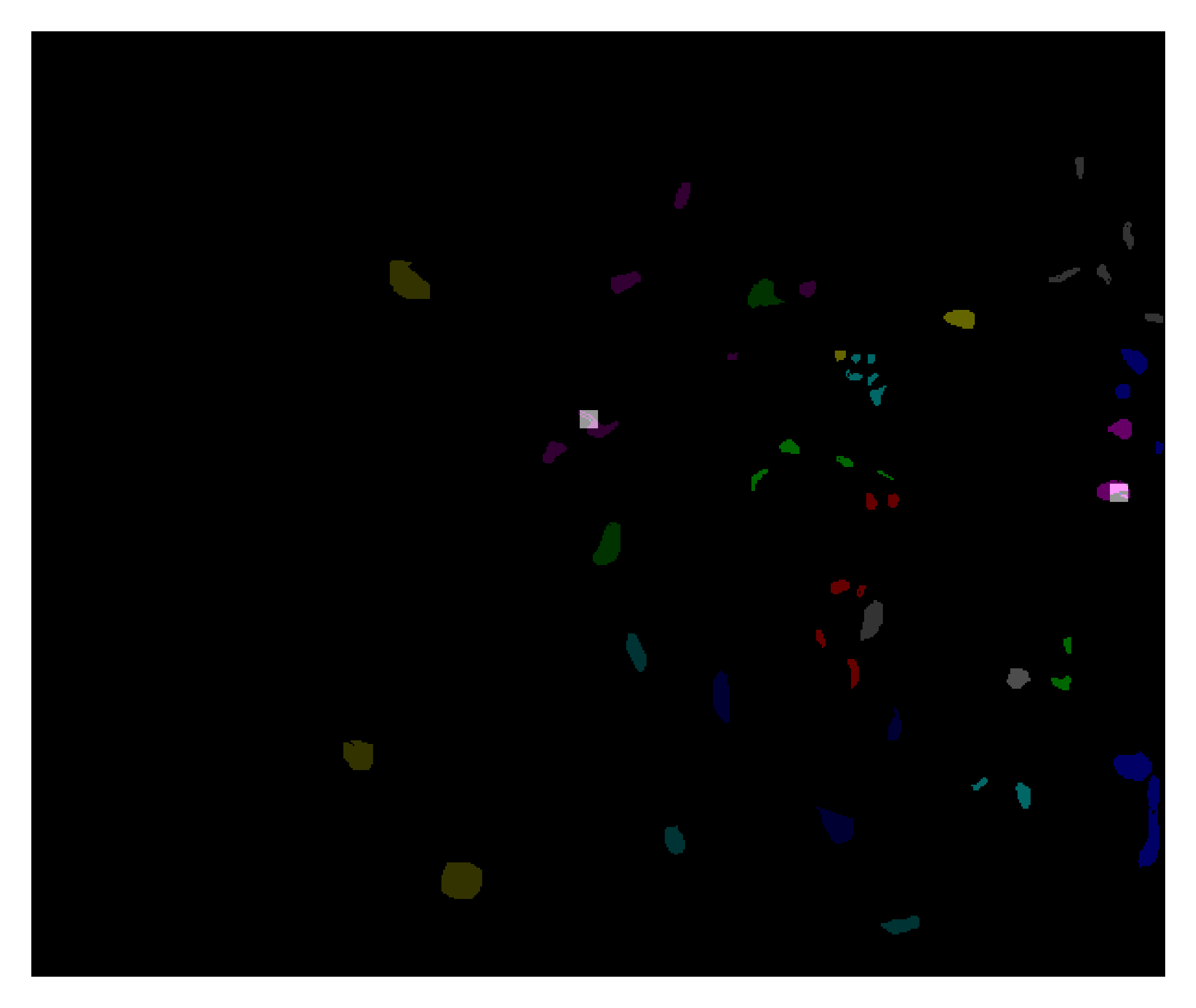}%
  \label{fig:R1_7_kscb}}
  \hfil
  \subfloat[SSRN]{\includegraphics[width=0.24\linewidth]{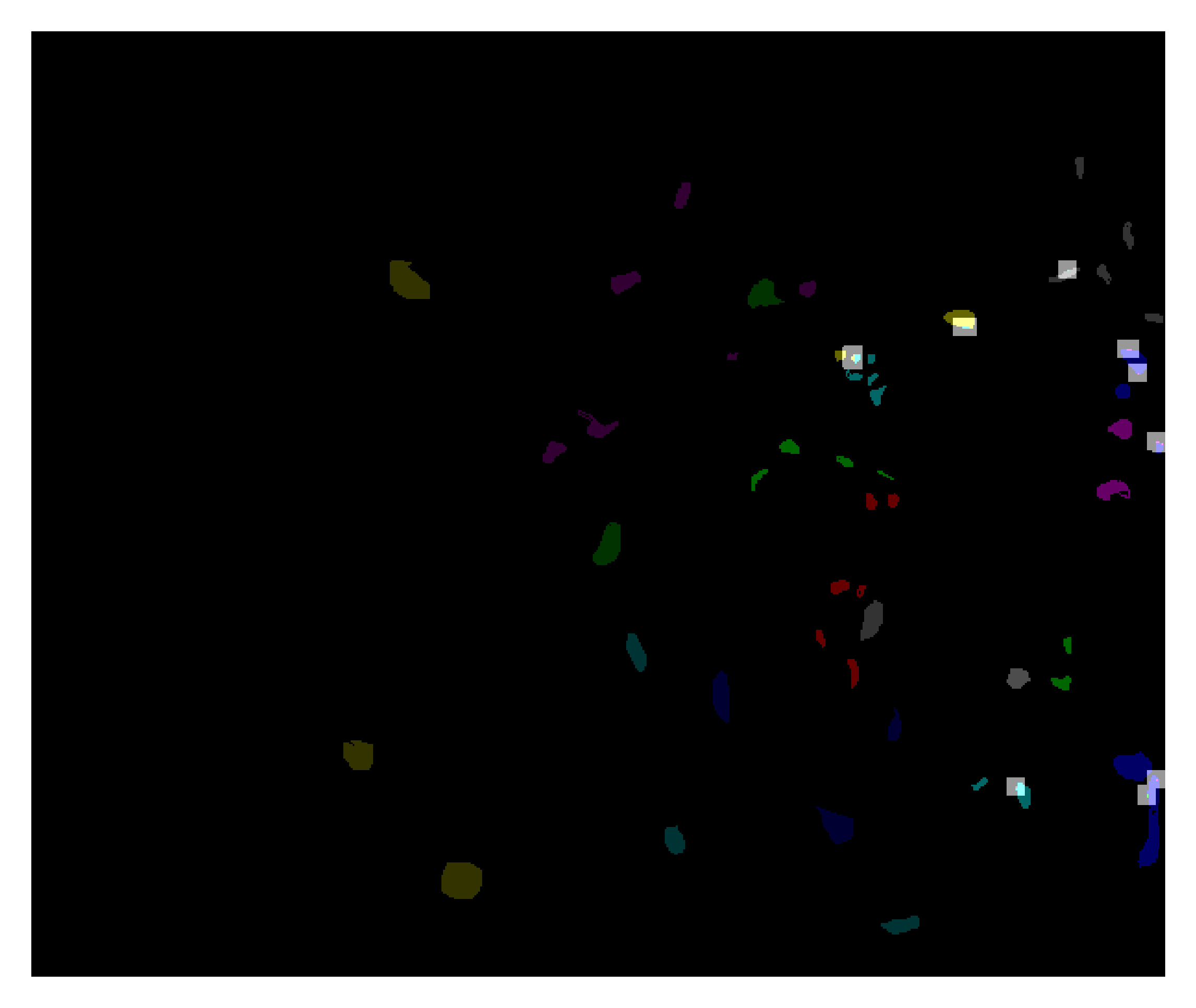}%
  \label{fig:R1_7_kscc}}
  \hfil
  \subfloat[A2S2KResNet]{\includegraphics[width=0.24\linewidth]{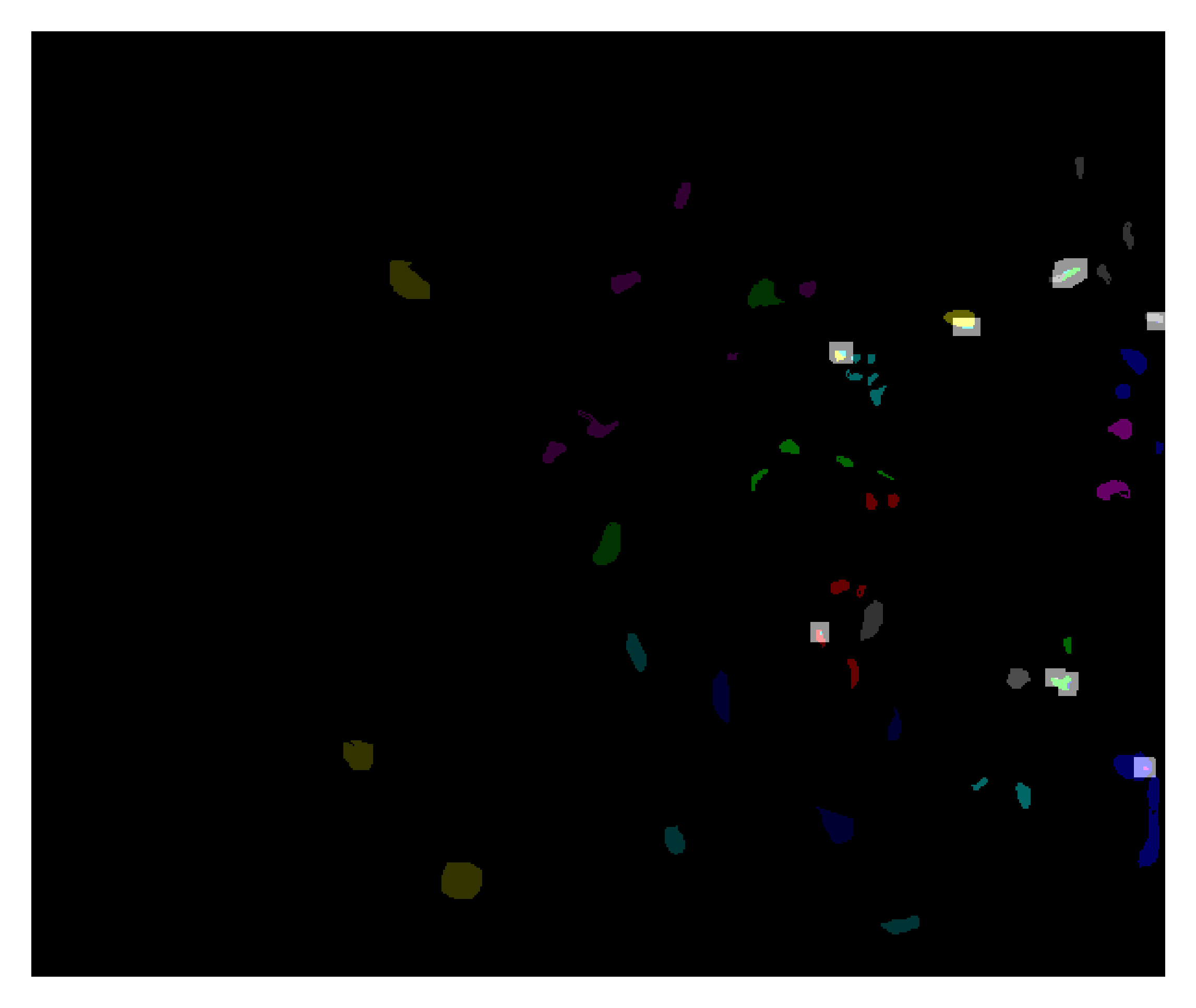}%
  \label{fig:R1_7_kscd}}
  \caption{Classification maps for Kennedy Space Centre (KSC) dataset.The main differences with respect to groundtruth have been highlighted. As one can observe, the number of errors of Triplet-Watershed is small compared to SSRN and A2S2K.}
  \label{fig:R1_7_ksc}
\end{figure*}

\begin{figure*}
  \tiny
  \centering
  % \subfloat[GT]{\includegraphics[width=0.24\linewidth]{./img/FigureR1_7/paviaU_gt.png}%
  % \label{fig:R1_7_paviaUa}}
  % \hfil
  % \subfloat[Triplet-Watershed]{\includegraphics[width=0.24\linewidth]{./img/FigureR1_7/paviaU_TripletWatershed.png}%
  % \label{fig:R1_7_paviaUb}}
  % \hfil
  % \subfloat[SSRN]{\includegraphics[width=0.24\linewidth]{./img/FigureR1_7/paviaU_SSRN.png}%
  % \label{fig:R1_7_paviaUc}}
  % \hfil
  % \subfloat[A2S2KResNet]{\includegraphics[width=0.24\linewidth]{./img/FigureR1_7/paviaU_A2S2K.png}%
  \subfloat[GT]{\includegraphics[width=0.24\linewidth]{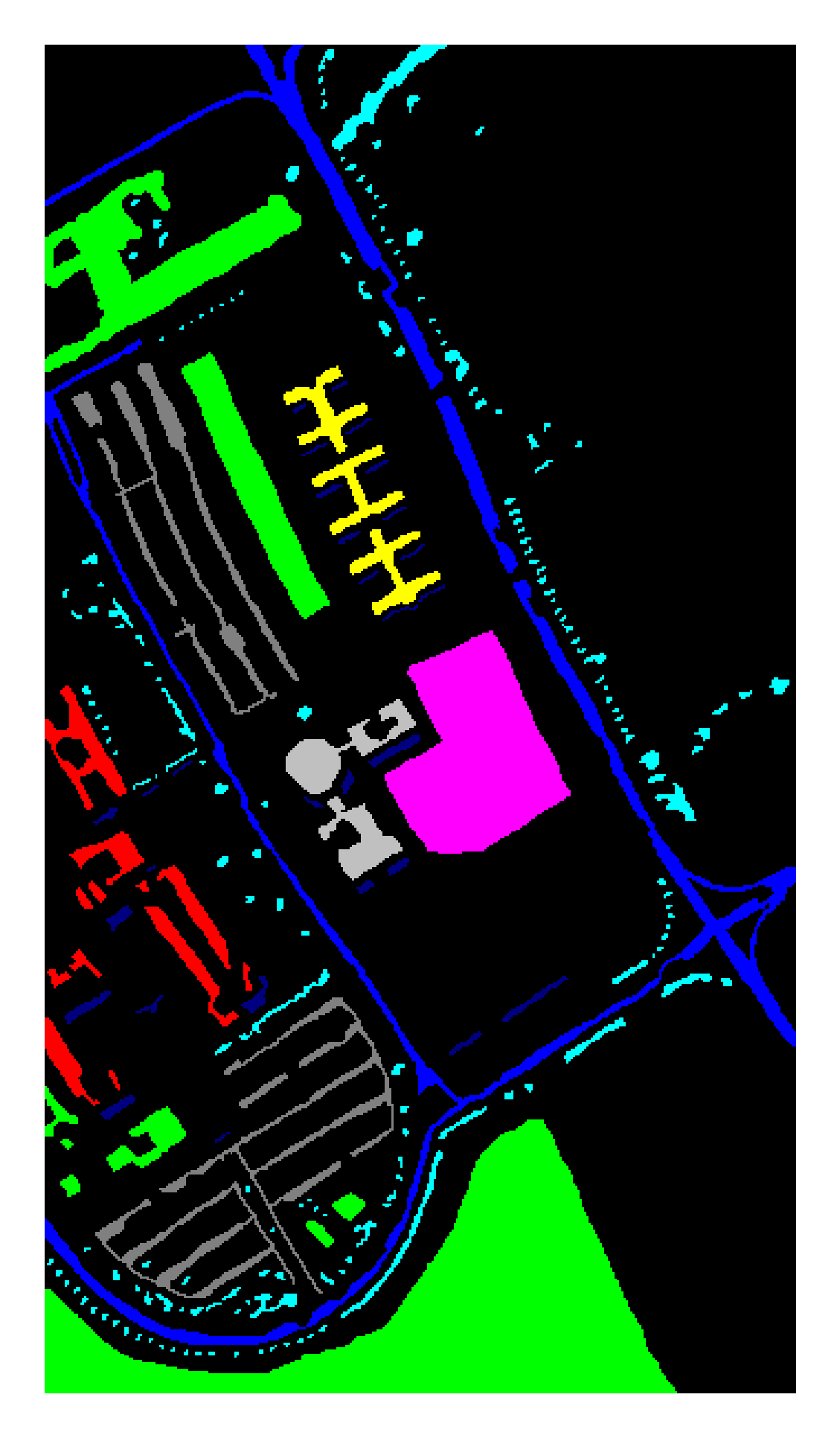}%
  \label{fig:R1_7_paviaUa}}
  \hfil
  \subfloat[Triplet-Watershed]{\includegraphics[width=0.24\linewidth]{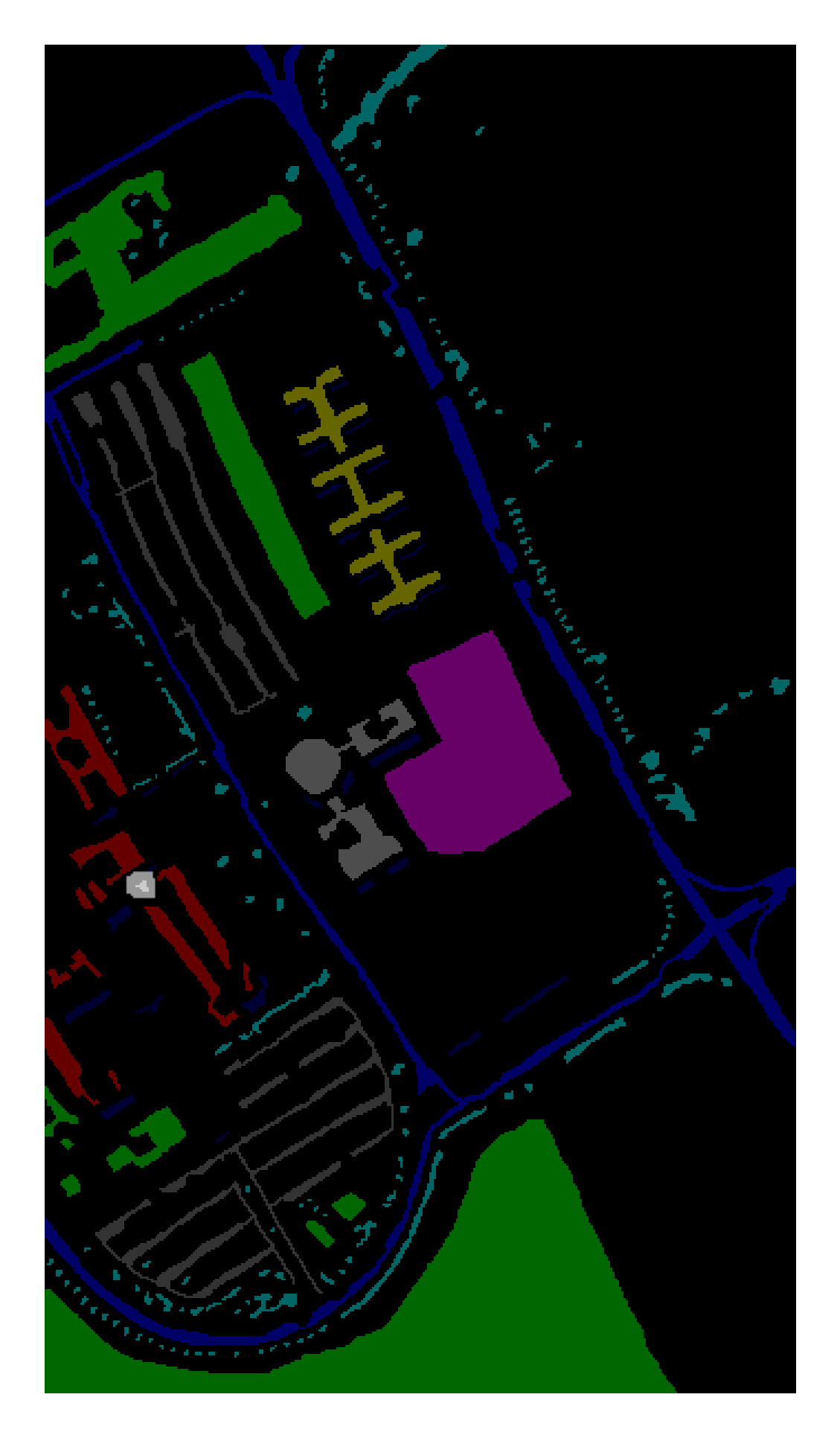}%
  \label{fig:R1_7_paviaUb}}
  \hfil
  \subfloat[SSRN]{\includegraphics[width=0.24\linewidth]{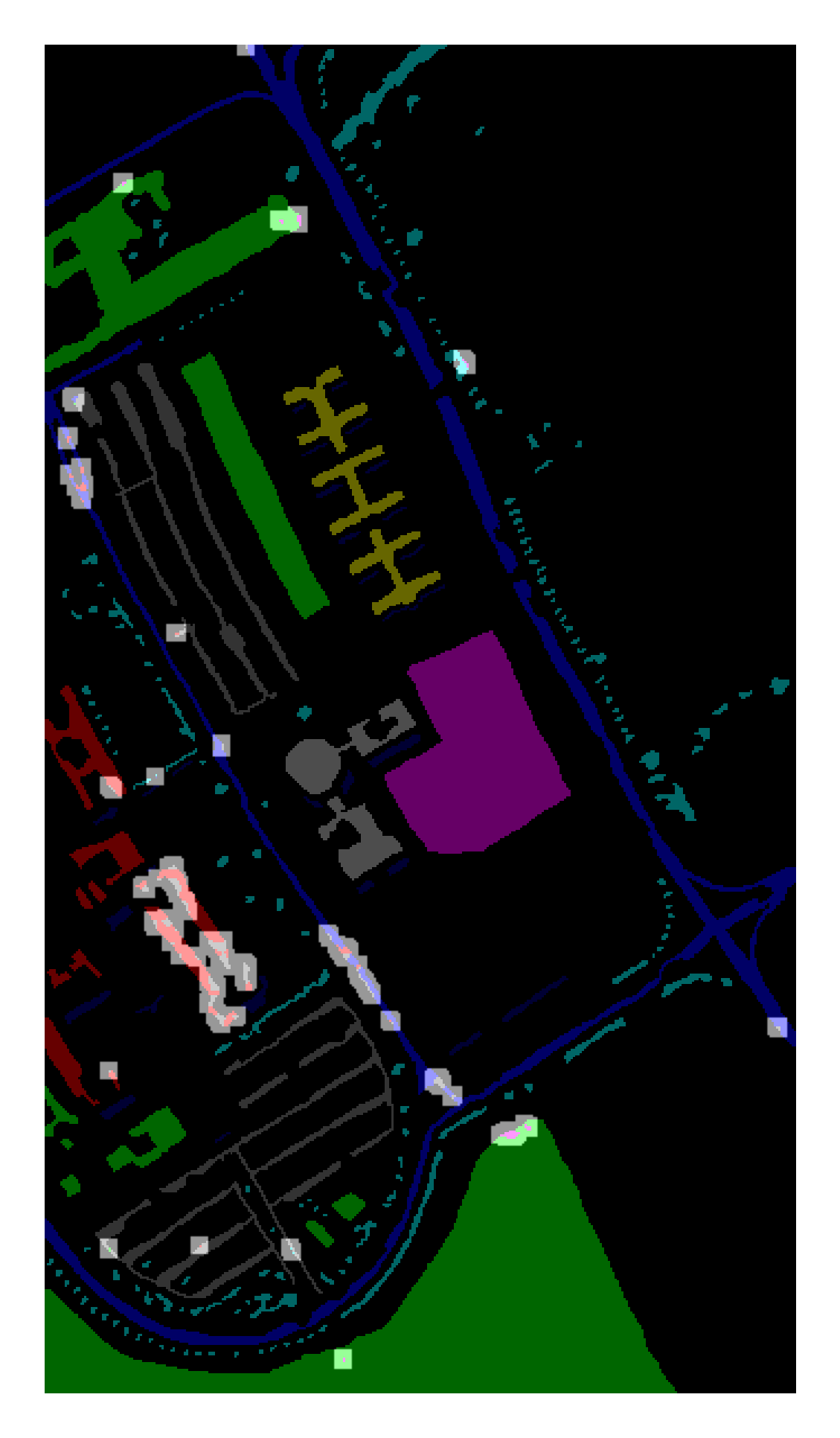}%
  \label{fig:R1_7_paviaUc}}
  \hfil
  \subfloat[A2S2KResNet]{\includegraphics[width=0.24\linewidth]{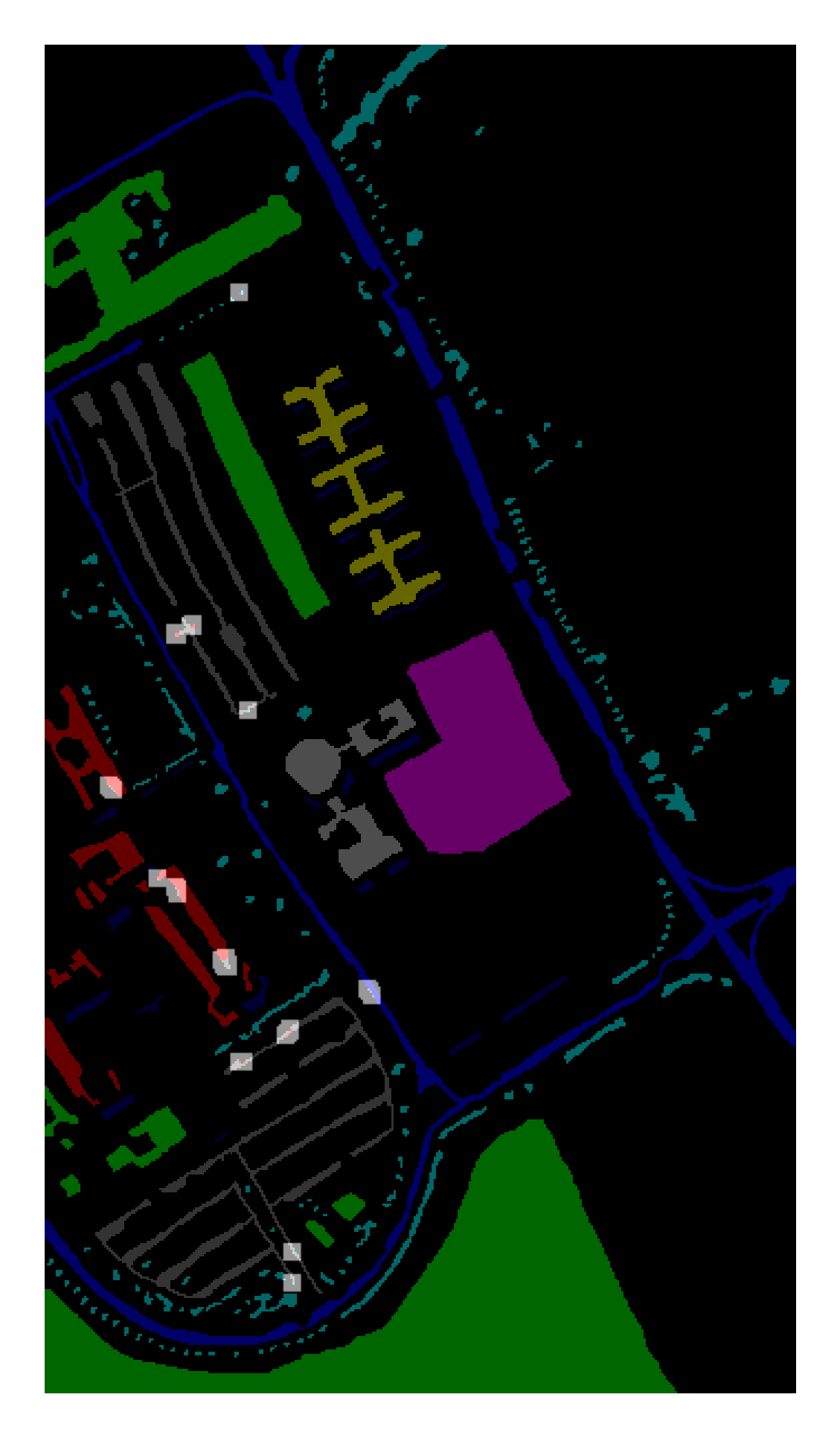}%
  \label{fig:R1_7_paviaUd}}
  \caption{Classification maps for University of Pavia (UP) dataset.The main differences with respect to groundtruth have been highlighted. As one can observe, the number of errors of Triplet-Watershed is small compared to SSRN and A2S2K.}
  \label{fig:R1_7_paviaU}
\end{figure*}

\begin{figure*}
  \tiny
  \centering
  % \subfloat[GT]{\includegraphics[width=0.4\linewidth]{./img/FigureR1_7/houston_gt.png}%
  % \label{fig:R1_7_houstona}}
  % \hfil
  % \subfloat[Triplet-Watershed]{\includegraphics[width=0.4\linewidth]{./img/FigureR1_7/houston_TripletWatershed.png}%
  % \label{fig:R1_7_houstonb}}
  
  % \subfloat[SSRN]{\includegraphics[width=0.4\linewidth]{./img/FigureR1_7/houston_SSRN.png}%
  % \label{fig:R1_7_houstonc}}
  % \hfil
  % \subfloat[A2S2KResNet]{\includegraphics[width=0.4\linewidth]{./img/FigureR1_7/houston_A2S2K.png}%
  % \label{fig:R1_7_houstond}}
  \subfloat[GT]{\includegraphics[width=0.4\linewidth]{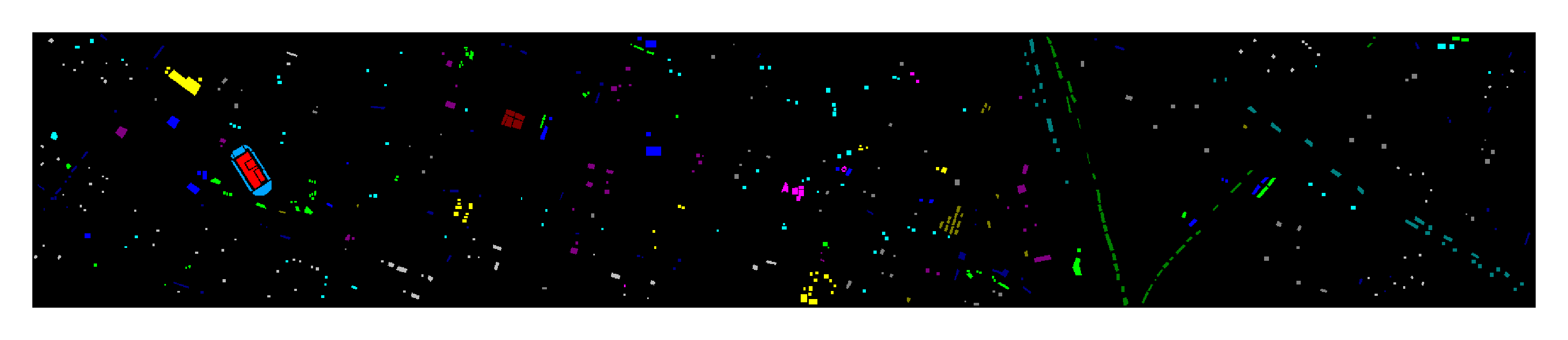}%
  \label{fig:R1_7_houstona}}
  \hfil
  \subfloat[Triplet-Watershed]{\includegraphics[width=0.4\linewidth]{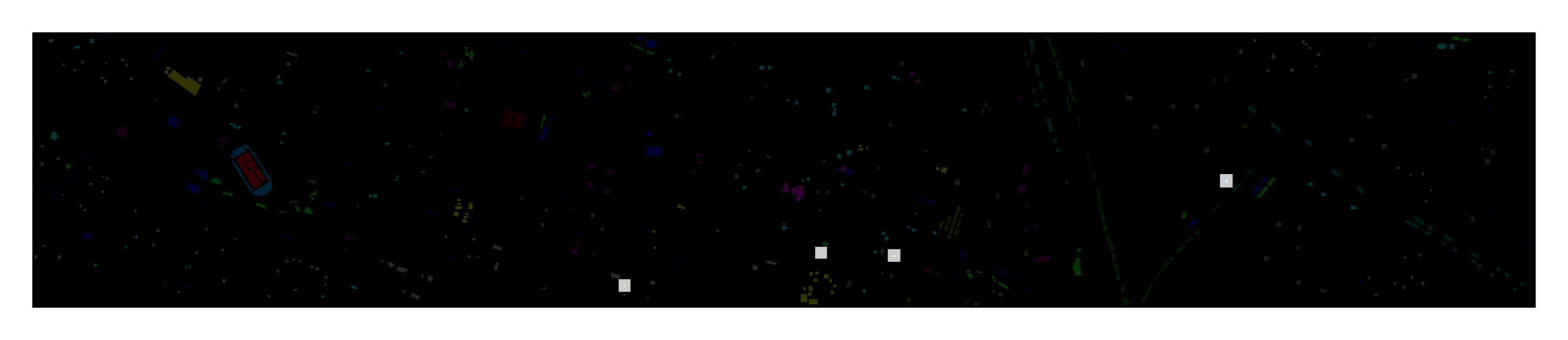}%
  \label{fig:R1_7_houstonb}}
  
  \subfloat[SSRN]{\includegraphics[width=0.4\linewidth]{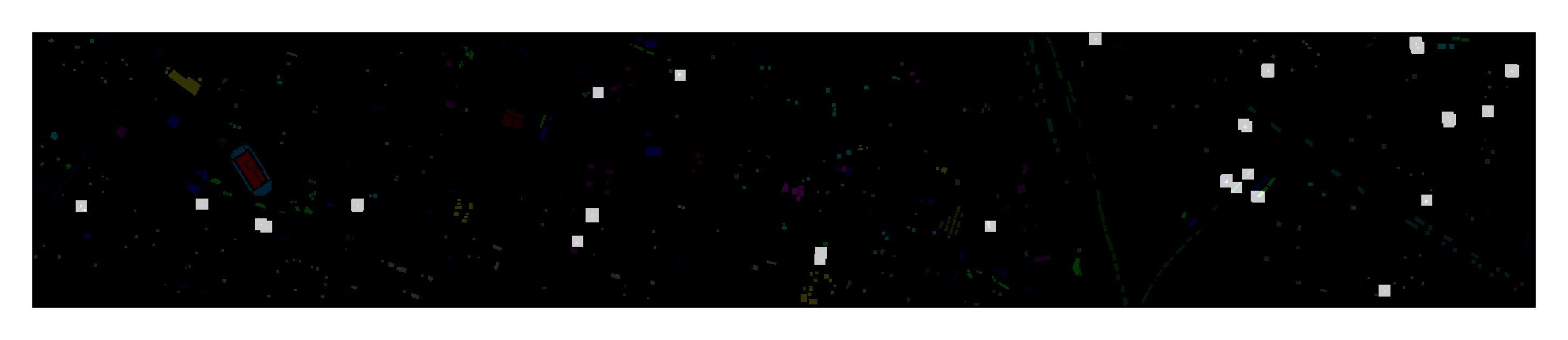}%
  \label{fig:R1_7_houstonc}}
  \hfil
  \subfloat[A2S2KResNet]{\includegraphics[width=0.4\linewidth]{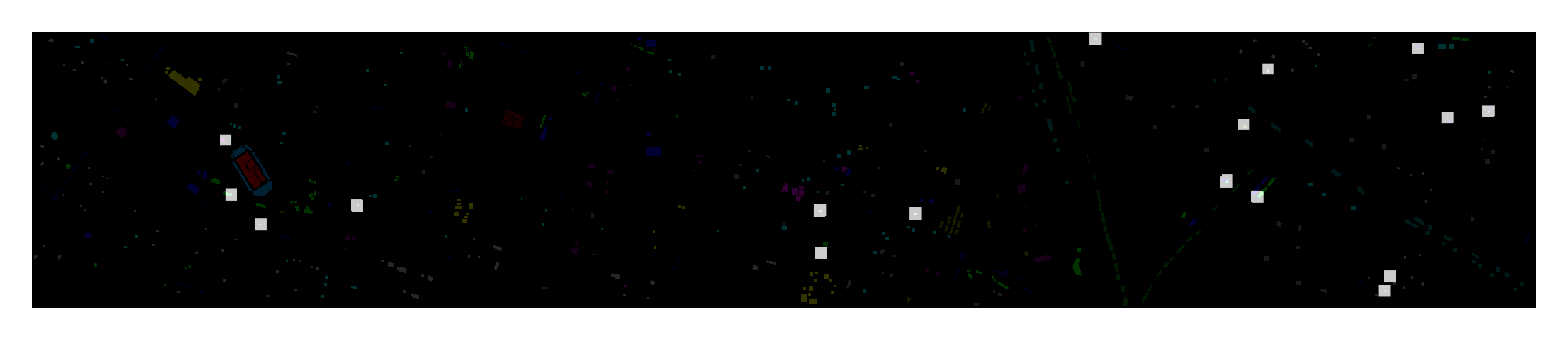}%
  \label{fig:R1_7_houstond}}
  \caption{Classification maps for University of Houston (UH) dataset.The main differences with respect to groundtruth have been highlighted. As one can observe, the number of errors of Triplet-Watershed is small compared to SSRN and A2S2K.}
  \label{fig:R1_7_houston}
\end{figure*}

\subsection{Supervised Classification}

Firstly, we provide the results of Triplet-Watershed for supervised classification. We compare our approach with standard baselines (SVM \cite{DBLP:journals/tgrs/MelganiB04} and Random Forest\cite{IEEE:journals/tgrs/Ham2005}), and also with the two recent state-of-art methods SSRN \cite{IEEE/journals/tgrs/Zhong2018} and A2S2K \cite{IEEE/journals/tgrs/Roy2020}. Tables \ref{table:2}, \ref{table:3}, \ref{table:4} show the results (OA, AA, $\kappa$) obtained. The train test splits per class are described in these tables. Note that Triplet-Watershed outperforms existing state-of-art A2S2KResNet\cite{IEEE/journals/tgrs/Roy2020} and other approaches in several aspects.  This can be attributed to the fact that - {Triplet Watershed exploits the connectivity patterns (edges within the pixels) in the dataset to propagate labels. Other approaches treat each pixel as a separate entity which would not exploit this observation}. Other approaches treat each pixel as a separate entity which would not exploit this observation. Simple Ensemble-Watershed results are shown in the tables as well. {Classification maps for Triplet-Watershed along with competing approaches are shown in figures \ref{fig:R1_7_indianpines},\ref{fig:R1_7_ksc},\ref{fig:R1_7_paviaU},\ref{fig:R1_7_houston}. { High resolution stand-alone images can also be found in \url{https://github.com/ac20/TripletWatershed_Code/tree/main/img/classification_maps}}}.

\subsection{Semi-Supervised Classification}

We compare the Triplet-Watershed with three recent state-of-art semi-supervised approaches - S2GCN\cite{IEEE:journal/grsl/Qin2019}, SSRN\cite{IEEE/journals/tgrs/Zhong2018} and DC-GCN (Dual Clustering GCN)\cite{ARXIV:Zeng2020}. We consider $30$ samples for training if the class size is greater than $30$ and $15$ if the class size is less than $30$. Tables \ref{table:2Semi}, \ref{table:3Semi} show the results obtained. Observe that, once again, Triplet-Watershed obtains the state-of-art in several aspects. 

\subsection{Evaluation of Representation}

Recall that accuracies in tables \ref{table:2}-\ref{table:3Semi} for Triplet-Watershed use ensemble watershed classifier. However, ensemble watershed exploits the connectivity patterns in the data. We now try to understand how well watershed representations compare with representations obtained by other approaches. Qualitatively, we use the TSNE\cite{GOOGLE:journal/jmlr/Maaten2008} plots as in Figure~\ref{fig:6}. Note that there does not exist any major differences except that within a class, A2S2K and SSRN have ``clumps'' points while Triplet-Watershed has a uniform density. Quantitatively we use the  mean average precision (MAP) over all points. The computation procedure is as follows:
\begin{enumerate}
  \item Given a data point $x_k$, we order all other data points $\{y_i\}_i$ using an inverse function of distance, $\exp(-\text{distance})$.
  \item Labels are assigned based on whether the points $\{y_i\}_i$ belong to the same class as $x_k$ or not with class label $1$ and $0$ respectively.
  \item Average precision (AP) computes the area under the precision-recall curve.
  \item The AP scores are averages over all points $\{x_k\}_k$ to obtain the MAP score.
\end{enumerate}
This procedure is as suggested in \cite{musgrave2020metric} to evaluate representations. The results are shown in Table \ref{table:6}. Observe that the watershed outperforms the current state-of-art techniques.

\subsection{Ablation Study}

We now study the importance of various aspects of Triplet-Watershed for the accuracies.

\subsubsection{Accuracy vs $\%$ training data}

Figure \ref{fig:5} shows the plots of overall accuracy (OA) vs $\%$ training data. For IP and UP datasets, it can be seen that Triplet-Watershed outperforms other approaches even at small sizes of training data. This can be attributed to the fact that the watershed classifier propagates the information to unlabelled nodes, which is in turn used for training. (See Figure \ref{fig:3b}). For optimal performance, the watershed classifier requires at least one labelled node per component. In cases of very small training data and many components, Triplet-Watershed does not perform well. This is the case for the KSC dataset at $2\%$ and $3\%$ training data, as shown in Figure \ref{fig:5}. { Detailed analysis of the underperformance of Triplet-Watershed at low train sizes for Kennedy Space Center (KSC) and University of Houston (UH) dataset can be found in appendix \ref{sec:appendix 2}.}

\subsubsection{Replacing Watershed With Other Classifiers}
\label{subsubsection:replace watershed}

{To illustrate the importance of the watershed classifier in the training pipeline (Figure \ref{fig:3b}), we replace it with Random Forest (RF) classifier and K-Nearest Neighbors (KNN) classifier with $k=5$, referring to these as {\em Triplet-Random Forest} and {\em Triplet-K-Nearest-Neighbors}. The results are shown in Table \ref{table:5}. Firstly observe the dramatic improvement of accuracies with respect to vanilla classifiers (Tables \ref{table:2}, \ref{table:3}, \ref{table:4}). Also, observe that Triplet-Watershed outperforms the other techniques. This, once again, is attributed to the fact that watershed exploits the observation that classes in the groundtruth consist of connected components. }

{
\noindent
\textbf{Remark}: Both Random Forest (RF) and K-Nearest Neighbors (KNN) are considered for this experiment since the labels generated by these are not differentiable with respect to the input representations. This property is shared with the watershed classifier. However, Multi-layered perceptron (MLP) and Support vector machines (SVM) obtain labels using specific costs and are indeed differentiable with respect to their input representations. Hence, the latter approaches are not considered for comparison. 
}

\subsubsection{Accuracy vs embed dimension}

Table \ref{table:7} shows the effect of embedding dimension on accuracy. Observe that there does not exist any significant trend with respect to the embedding dimension. We use $64$ as the default embedding dimension. 

{
\subsubsection{Accuracy Vs Patch Size}

Recall that one of the hyperparameter of the approach is patch size - The size of the window around the pixel. Table \ref{table:R2_9} shows the results obtained by varying the patch sizes across different datasets. Observe that larger window size implies more information for inference and hence scope for better inference. Thus, as a rule of thumb, larger window size obtain better results. But, it also implies higher computational requirement. However in several cases increasing the window size beyond a threshold would not lead to significant improvements. For example, in table \ref{table:R2_9} IN and UP datasets do not show much improvement with larger window sizes. UH dataset improves with larger window size, but no significant improvement is obtained by increasing the window size from $11$ to $13$. }

\section{Conclusion}
In this article, we proposed a novel approach to train for the watershed classifier. We refer to this as Triplet-Watershed. We show that the watershed classifier exploits the connectivity patterns in the datasets. This leads to huge performance gains compared to other approaches which use simple softmax classifier. We prove this empirically by comparing Triplet-Watershed with existing state-of-art deep learning approaches such as A2S2K\cite{IEEE/journals/tgrs/Roy2020}, SSRN\cite{IEEE/journals/tgrs/Zhong2018} and also classic approaches - RF\cite{IEEE:journals/tgrs/Ham2005} and SVM\cite{DBLP:journals/tgrs/MelganiB04}. We also compare the current technique with semi-supervised approaches such as  S2GCN\cite{IEEE:journal/grsl/Qin2019} and DC-GCN\cite{ARXIV:Zeng2020}. In each case, we achieve better accuracy while using a quarter of the parameters of the previous state-of-the-art approaches.

\appendices

\section{Constructing the graph on HSI}
\label{sec:appendix 1}
{
\begin{figure}
  
  \subfloat[Groundtruth]{\includegraphics[width=0.4\linewidth]{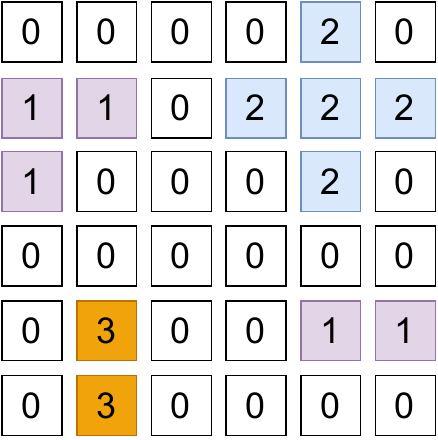}%
  \label{fig:A1a}}
  \hfil
  \subfloat[Graph]{\includegraphics[width=0.4\linewidth]{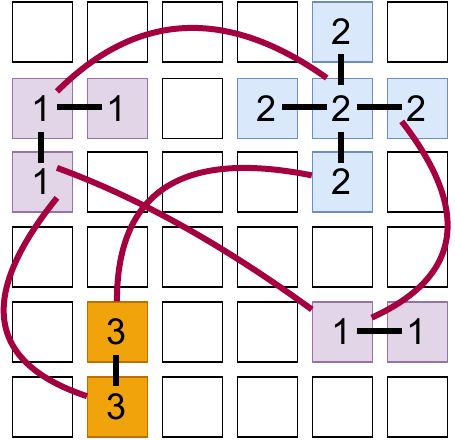}%
  \label{fig:A1b}}
  \caption{Constructing the graph on HSI data. (a) shows a simple toy HSI data with groundtruth classes. Note that class $0$ implies that groundtruth is not available. (b) illustrates the graph constructed by considering only points with $\{\text{labels} \neq 0\}$ as vertices. 4-adjacency edges (black) along with other edges (red) spanning across components are considered. These ``other'' edges are constructed using techniques such as Euclidean Minimum Spanning Tree (EMST) or K-Neighbors graph.}
  \label{fig:A1}
\end{figure}

Here, we illustrate the process of constructing the graph on HSI dataset. Figure \ref{fig:A1a} considers a simple hypothetical image with the groundtruth classes as shown. Figure \ref{fig:A1b} shows the graph obtained using the following steps:
\begin{enumerate}[label=(\roman*)]
  \item Firstly, only points with groundtruth available, i.e $\{\text{labels} \neq 0\}$ are considered. This can be trivially extended to other points depending on requirement. These points constitute the vertex set.
  \item The edge set is obtained by taking the union of - (a) 4 adjacency edges denoted by colour black and (b) ``other'' edges which span across components. These ``other'' edges are constructed using Euclidean Minimum Spanning Tree (EMST) for IP, UP, and KSC datasets. For UH dataset these edges are constructed using K-Neighbors graph with k=50. 
\end{enumerate}

{The two main principles for selecting the graph are - (i) We require each label-induced subgraph\footnote{\label{footnote1}Given a graph $G = (V, E, W)$, the subgraph induces by a subset of vertices $V' \subset V$ is given by $G' = (V', E', W)$. Here $E' = \{(e_x, e_y) \in E \text{ such that } e_x, e_y \in V'\}$ }such that the number of connected components are as few as possible and (ii) We also require the number of edges to be as few as possible. Both these act against each other and the right combination is obtained through trial and error. }
}
 
\section{Triplet-Watershed at small train sizes}
\label{sec:appendix 2}

\begin{table*}

\caption{Sizes of components of label-induced subgraphs for datasets UH and IP. Three kinds of labels are considered - Groundtruth, labels predicted at $2\%$ and $10\%$. Also, shown are the relative (to maximum) standard deviations of the groundtruth components.}
\label{table:A1}
\centering
\begin{adjustbox}{max width=.95\linewidth}
\begin{tabular}{ccccccccccc}
    \toprule
    \multicolumn{5}{c}{UH} & \phantom{abc} & \multicolumn{5}{c}{IP}\\
    Label & Groundtruth & Rel. Stdev. &10\%  & 2\% & \phantom{abc} &Label & Groundtruth & Rel. Stdev. &10\%  & 2\% \\
    \midrule
1  & [178, 1073]            &[0.41, 0.68]  &[154, 1073]         & [1073]  & \phantom{abc}                          & 1  & [46]           & [0.79] & [46]            & [46]  \\
\rowcolor{Gray}2  & [1096, 158]            &[0.68, 0.4]  &[1096, 158]         & [312, 1130]  & \phantom{abc}                     & 2  & [1428]          & [0.76] & [1416, 1, 1]    & [1440]  \\
3  & [697]                  &[0.47]  &[697]               & [697]  & \phantom{abc}                                 & 3  & [830]          & [0.73] & [830]           & [830]  \\
\rowcolor{Gray}4  & [1174, 70]             &[0.61, 0.39]  &[1268]              & [1268]  & \phantom{abc}                               & 4  & [237]     & [0.83] & [237]           & [237]  \\
5  & [1242]                 &[0.72]  &[1272]              & [1315]  & \phantom{abc}                               & 5  & [318,147,18]    & [0.65, 0.74, 0.72] & [318, 147, 18]  & [318, 88]  \\
\rowcolor{Gray}6  & [40, 6, 279]           &[0.51, 0.64, 0.65]  &[40, 6, 279]        &[6, 279]   & \phantom{abc}                       & 6  & [730]     & [0.68] & [730]           & [730]  \\
7  & [1268]                 &[0.56]  &[1268]              & [1225]  & \phantom{abc}                                     & 7  & [28]      & [0.68] & [28]            & [28]  \\
\rowcolor{Gray}8  & [1011, 170, 34, 20, 9] &[0.93, 0.42, 0.66, 0.56, 0.63]  &[998, 170, 34, 20]  & [290, 224, 476] & \phantom{abc} & 8  & [478]         & [0.85] & [478]           & [478]  \\
9  & [1243, 9]              &[0.66, 0.64]  &[1257, 9]           & [1302, 9, 1]  & \phantom{abc}                         & 9  & [20]      & [0.67] & [20]            & [20]  \\
\rowcolor{Gray}10 & [901, 326]             &[0.65, 0.42]  &[905, 326]          & [908, 368]  & \phantom{abc}                         & 10 & [912, 60]   & [0.69, 0.84] & [912, 60]       & [912, 60]  \\
11 & [1235]                 &[0.56]  &[1204]              & [1176, 118]  & \phantom{abc}                                & 11 & [2455]    & [0.74] & [2465]          & [2520]  \\
\rowcolor{Gray}12 & [1233]                 &[0.65]  &[1237]              & [1726]  & \phantom{abc}                                     & 12 & [593]     & [0.82] & [633]           & [624]  \\
13 & [469]                  &[1.0]  &[461]               & [3, 13, 1, 1]  & \phantom{abc}                                & 13 & [205]    & [0.68] & [205]           & [205]  \\
\rowcolor{Gray}14 & [428]                  &[0.95]  &[428]               & [428]  & \phantom{abc}                                        & 14 & [1265]  & [0.74] & [1265]          & [1265]  \\
15 & [660]                  &[0.67]  &[669]               & [680]  & \phantom{abc}                                        & 15 & [386]   & [0.76] & [386]           & [386]  \\
\rowcolor{Gray}\phantom{abc} &\phantom{abc} &\phantom{abc} &\phantom{abc} &\phantom{abc} & \phantom{abc}          & 16 & [93]                           & [1.0] & [53]            & [62]  \\
\bottomrule
\end{tabular}
\end{adjustbox}
\end{table*}

{
Note that from figure \ref{fig:5}, at low train sizes ($2\%$ and $3\%$, Triplet-Watershed performs better than A2S2KResNet and SSRN on IP, UP datasets. While, Triplet-Watershed is slightly inferior to A2S2KResNet and SSRN on KSC, UH datasets. In this section we analyze and explain this in detail.

There are two main reasons for the different behaviours of Triplet-Watershed at high ($10\%$) and low ($2\%$, $3\%$) train sizes - (i) At low train sizes, not all components within the data are covered and (ii) There aren't enough points near the boundary to allow for better separation. To understand this better, we perform a post-hoc analysis on UH and IP datasets. 

For each label, (both groundtruth and prediction) we consider the subgraph induced by the vertices\footnote{See footnote \ref{footnote1}.} of the given label. In this subgraph, we count the size of each connected component. Table \ref{table:A1} shows these values for UH/IP datasets, for groundtruth labels, and labels predicted for $10\%$ and $2\%$. Both the above phenomenon can be observed in table \ref{table:A1}.
\begin{enumerate}[label=(\roman*)]
  \item Observe that for several classes in UH dataset, there exists small components for UH (example : class 1 with 178 points) which are not represented when only $2\%$ of the data is considered for training. While, this happens for IP dataset (class 5, 147 points), it is relatively low in magnitude. This partly explains why we achieve better results at $10\%$ train size. And also why IP performs better at $2\%$ train size comparatively.
  \item The other main reason is - Boundaries are not sufficiently represented at $2\%$ train size. As an example of this, consider class 13 for UH dataset which has a single component $469$ points. At $2\%$ train size, this component splits into small components. However, at $10\%$ train size, the component is preserved. This is due to insufficient boundary information at $2\%$ train size. Moreover, as can be intuitively expected, this happens when there is a relatively high standard deviation within the class. 
\end{enumerate}

The above observations explain the behaviour of Triplet-Watershed at low train sizes. 

}

  \section*{Acknowledgment}

  All the authors would like to thank the Associate-Editor, Editor-in-Chief and the anonymous reviewers for their valuable comments. AC would like to thank Indian Institute of Science for the Raman Fellowship and BITS-Pilani K K Birla Goa Campus for the support. SD would like to acknowledge the funding received from BPGC/RIG/2020-21/11-2020/01 (Research Initiation Grant provided by BITS-Pilani K K Birla Goa Campus). The work of B. S. D. Sagar was supported by the DST-ITPAR-Phase-IV project and the Technology Innovation Hub on Data Science, Big Data Analytics and Data Curation project sanctioned under the National Mission for the Interdisciplinary Cyber-Physical Systems respectively under the Grant numbers INT/Italy/ITPAR-IV/Telecommunication/2018, and NMICPS/006/MD/2020-21. The work of Laurent Najman is supported by Programme d'Investissements d'Avenir (LabEx BEZOUT ANR-10-LABX-58). 

\ifCLASSOPTIONcaptionsoff
  \newpage
\fi



\bibliographystyle{IEEEtran}
\bibliography{references}

\begin{IEEEbiography}[{\includegraphics[width=1in,height=1.25in,clip,keepaspectratio]{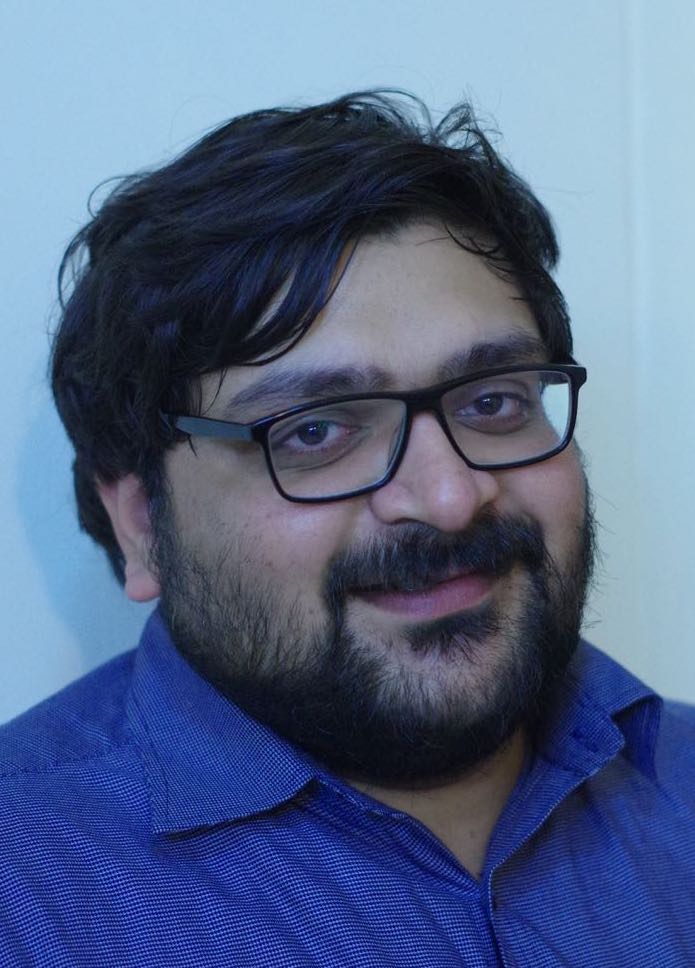}}]{Aditya Challa}
    received the B.Math.(Hons.) degree in Mathematics from the Indian Statistical Institute - Bangalore, and  Masters in Complex Systems from University of Warwick, UK - in 2010, and 2012, respectively.  From 2012 to 2014, he worked as a Business Analyst at Tata Consultancy Services, Bangalore. He completed his PhD in computer science from Systems Science and Informatics Unit, Indian Statistical Institute - Bangalore. He is currently Raman PostDoc Fellow at Indian Institute of Science, Bangalore. His current research interests focus on using techniques from Mathematical Morphology in Machine Learning.
  \end{IEEEbiography}
  
  \begin{IEEEbiography}[{\includegraphics[width=1in,height=1.25in,clip,keepaspectratio]{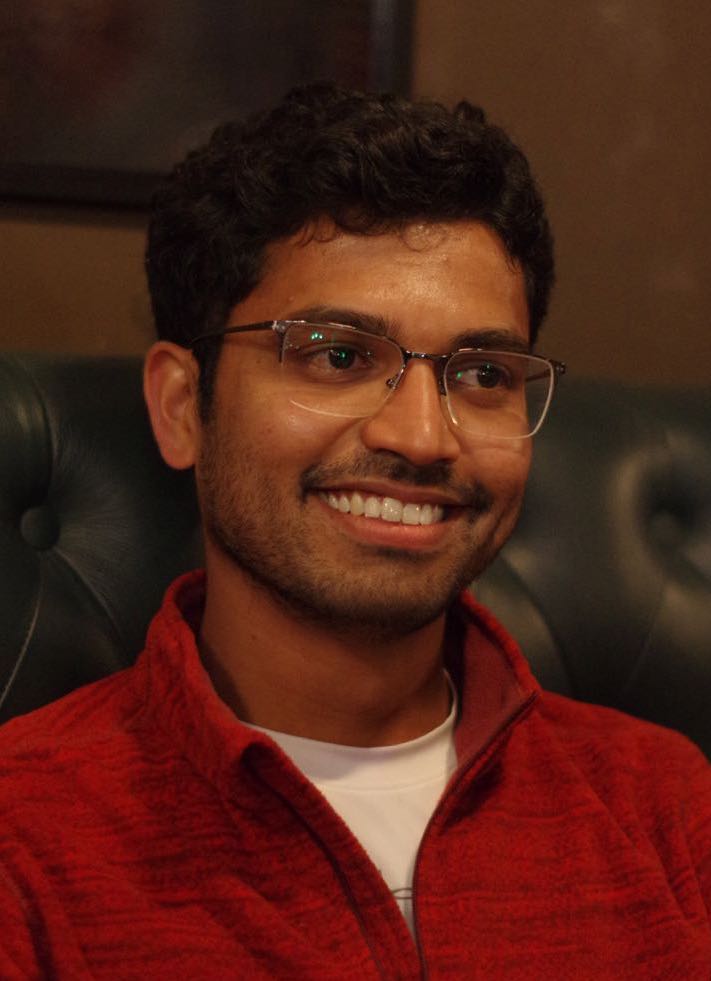}}]{Sravan Danda}
    received the B.Math.(Hons.) degree in Mathematics from the Indian Statistical Institute - Bangalore, and the M.Stat. degree in Mathematical Statistics from the Indian Statistical Institute - Kolkata, in 2009, and 2011, respectively.  From 2011 to 2013, he worked as a Business Analyst at Genpact - Retail Analytics, Bangalore. He completed his PhD in computer science from Systems Science and Informatics Unit, Indian Statistical Institute - Bangalore under the joint supervision of B.S.Daya Sagar and Laurent Najman. He is currently working as a Assistant Professor at Department of Computer Science and Information Systems, BITS Pilani K K Birla Goa Campus. His current research interests are discrete mathematical morphology and discrete optimization.
  \end{IEEEbiography}
  
  \begin{IEEEbiography}[{\includegraphics[width=1in,height=1.25in,clip,keepaspectratio]{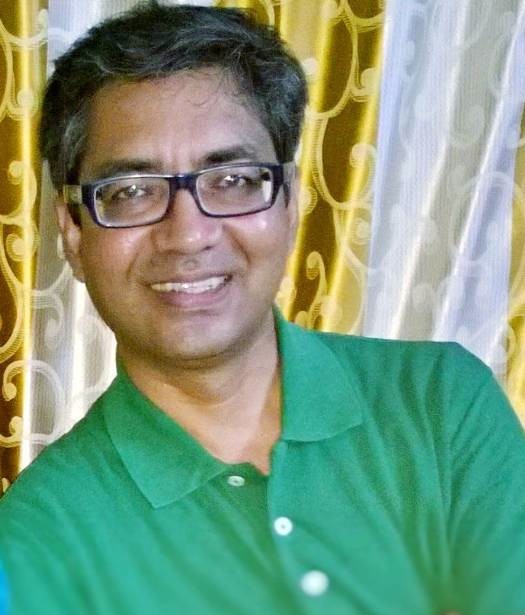}}]{B. S. Daya Sagar}
    (M’03-SM’03) is a Full Professor of the Systems Science and Informatics Unit (SSIU) at the Indian Statistical Institute. Sagar received his MSc and Ph.D. degrees in Geoengineering and Remote Sensing from the Faculty of Engineering, Andhra University, Visakhapatnam, India, in 1991 and 1994 respectively. He is also the first Head of the SSIU. Earlier, he worked in the College of Engineering, Andhra University, and Centre for Remote Imaging Sensing and Processing (CRISP), The National University of Singapore in various positions during 1992-2001. He served as Associate Professor and Researcher in the Faculty of Engineering \& Technology (FET), Multimedia University, Malaysia, during 2001-2007. Sagar has made significant contributions to the field of geosciences, with special emphasis on the development of spatial algorithms meant for geo-pattern retrieval, analysis, reasoning, modeling, and visualization by using concepts of mathematical morphology and fractal geometry. He has published over 85 papers in journals and has authored and/or guest-edited 11 books and/or special theme issues for journals. He recently authored a book entitled "Mathematical Morphology in Geomorphology and GISci," CRC Press: Boca Raton, 2013, p. 546. He recently co-edited two special issues on "Filtering and Segmentation with Mathematical Morphology" for IEEE Journal of Selected Topics in Signal Processing (v. 6, no. 7, p. 737-886, 2012), and "Applied Earth Observation and Remote Sensing in India" for IEEE Journal of Selected Topics in Applied Earth Observation and Remote Sensing (v. 10, no. 12, p. 5149-5328, 2017). His recent book “Handbook of Mathematical Geosciences”, Springer Publishers, p. 942, 2018 reached 750000 downloads. He was elected as a member of the New York Academy of Sciences in 1995, as a Fellow of the Royal Geographical Society in 2000, as a Senior Member of the IEEE Geoscience and Remote Sensing Society in 2003, as a Fellow of the Indian Geophysical Union in 2011. He is also a member of the American Geophysical Union since 2004, and a life member of the International Association for Mathematical Geosciences (IAMG). He delivered the "Curzon \& Co - Seshachalam Lecture - 2009" at Sarada Ranganathan Endowment Lectures (SRELS), Bangalore, and the "Frank Harary Endowment Lecture - 2019" at International Conference on Discrete Mathematics - 2019 (ICDM - 2019). He was awarded the 'Dr. Balakrishna Memorial Award’ of the Andhra Pradesh Academy of Sciences in 1995, the Krishnan Medal of the Indian Geophysical Union in 2002, the 'Georges Matheron Award - 2011 with Lectureship' of the IAMG, and the Award of IAMG Certificate of Appreciation - 2018. He is the Founding Chairman of the Bangalore Section IEEE GRSS Chapter. He has been recently appointed as an IEEE Geoscience and Remote Sensing Society (GRSS) Distinguished Lecturer (DL) for a two-year period from 2020 to 2022. He is on the Editorial Boards of Computers \& Geosciences, Frontiers: Environmental Informatics, and Mathematical Geosciences. He is also the Editor-In-Chief of the Springer Publishers’ Encyclopedia of Mathematical Geosciences. 
  \end{IEEEbiography}

  \begin{IEEEbiography}[{\includegraphics[width=1in,height=1.25in,clip,keepaspectratio]{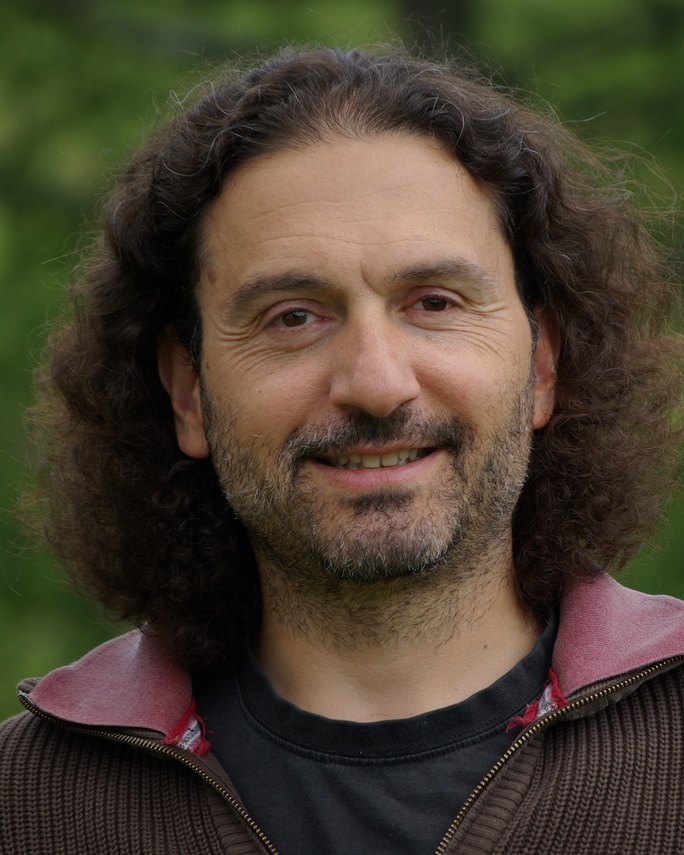}}]{Laurent Najman} (SM’17) received the Habilitation \`a Diriger les Recherches in 2006 from University the University of Marne-la-Vall\'ee, a Ph.D. of applied mathematics from Paris-Dauphine University in 1994 with the highest honor (Félicitations du Jury) and an ``Ing\'enieur'' degree from the Ecole des Mines de Paris in 1991. After earning his engineering degree, he worked in the central research laboratories of Thomson-CSF for three years, working on some problems of infrared image segmentation using mathematical morphology. He then joined a start-up company named Animation Science in 1995, as director of research and development. The technology of particle systems for computer graphics and scientific visualization, developed by the company under his technical leadership received several awards, including the ``European Information Technology Prize 1997'' awarded by the European Commission (Esprit programme) and by the European Council for Applied Science and Engineering and the ``Hottest Products of the Year 1996'' awarded by the Computer Graphics World journal. In 1998, he joined OC\'E Print Logic Technologies, as senior scientist. He worked there on various problem of image analysis dedicated to scanning and printing. In 2002, he joined the Informatics Department of ESIEE, Paris, where he is professor and a member of the Institut Gaspard Monge, Universit\'e Gustave Eiffel. His current research interest is discrete mathematical morphology and discrete optimization.
  
  \end{IEEEbiography}

\end{document}